\documentclass[sigconf]{acmart}
\AtBeginDocument{%
  \providecommand\BibTeX{{%
    \normalfont B\kern-0.5em{\scshape i\kern-0.25em b}\kern-0.8em\TeX}}}

\usepackage{url} 
\usepackage{rotating} 
\usepackage[ruled,vlined]{algorithm2e}
\usepackage{color, colortbl}
\usepackage{amsthm}
\usepackage{amsmath}
\usepackage{amsfonts}
\usepackage{caption}
\usepackage{multirow}
\usepackage{rotating}
\usepackage{comment}
\usepackage{enumerate}
\usepackage{dblfloatfix}
\usepackage{arydshln}

\makeatletter
\def\adl@drawiv#1#2#3{%
        \hskip.5\tabcolsep
        \xleaders#3{#2.5\@tempdimb #1{1}#2.5\@tempdimb}%
                #2\z@ plus1fil minus1fil\relax
        \hskip.5\tabcolsep}
\newcommand{\cdashlinelr}[1]{%
  \noalign{\vskip\aboverulesep
           \global\let\@dashdrawstore\adl@draw
           \global\let\adl@draw\adl@drawiv}
  \cdashline{#1}
  \noalign{\global\let\adl@draw\@dashdrawstore
           \vskip\belowrulesep}}
\makeatother

\newtheorem{lemma}{Lemma}
\newtheorem{lemmaproof}{Lemma}
\newtheorem{definition}{Definition}
\newtheorem{proposition}{Proposition}
\theoremstyle{definition}

\definecolor{Cyan}{rgb}{0.88,1,1}
\definecolor{Pink}{rgb}{1,0.75,0.80}
\definecolor{greenmatplotlib}{rgb}{0.0, 0.5019607843137255, 0.0}

\AtBeginDocument{%
  \providecommand\BibTeX{{%
    \normalfont B\kern-0.5em{\scshape i\kern-0.25em b}\kern-0.8em\TeX}}}

\copyrightyear{2020}
\acmYear{2020}
\setcopyright{acmlicensed}\acmConference[FODS '20]{Proceedings of the 2020
ACM-IMS Foundations of Data Science Conference}{October 19--20,
2020}{Virtual Event, USA}
\acmBooktitle{Proceedings of the 2020 ACM-IMS Foundations of Data Science
Conference (FODS '20), October 19--20, 2020, Virtual Event, USA}
\acmPrice{15.00}
\acmDOI{10.1145/3412815.3416893}
\acmISBN{978-1-4503-8103-1/20/10}

\begin{document}
\fancyhead{}

\title[Tree Space Prototypes: Another Look at Making Tree Ensembles Interpretable]{Tree Space Prototypes:\\ Another Look at Making Tree Ensembles Interpretable}

\author{Sarah Tan}
\email{ht395@cornell.edu}
\affiliation{
  \institution{Cornell University}
}

\author{Matvey Soloviev}
\email{ms2837@cornell.edu}
\affiliation{
  \institution{Cornell University}
}

\author{Giles Hooker}
\email{gjh27@cornell.edu}
\affiliation{
  \institution{Cornell University}
}

\author{Martin T. Wells}
\email{mtw1@cornell.edu}
\affiliation{
  \institution{Cornell University}
}

\renewcommand{\shortauthors}{Tan and Soloviev, et al.}

\begin{abstract}
Ensembles of decision trees perform well on many problems, but are not interpretable. In contrast to existing approaches in interpretability that focus on explaining relationships between features and predictions, we propose an alternative approach to interpret tree ensemble classifiers by surfacing representative points for each class -- prototypes. We introduce a new distance for Gradient Boosted Tree models, and propose new, adaptive prototype selection methods with theoretical guarantees, with the flexibility to choose a different number of prototypes in each class. We demonstrate our methods on random forests and gradient boosted trees, showing that the prototypes can perform as well as or even better than the original tree ensemble when used as a nearest-prototype classifier. In a user study, humans were better at predicting the output of a tree ensemble classifier when using prototypes than when using Shapley values, a popular feature attribution method. Hence, prototypes present a viable alternative to feature-based explanations for tree ensembles.
\end{abstract}



\maketitle

\section{Introduction}

As machine learning is increasingly employed alongside human reasoning in a wide range of tasks, it has been recognized that it is desirable for these systems to be made interpretable: a human user working alongside an ML system should be able to maintain a mental model of why and how the system arrives at its outputs, so she may either obtain confidence in the outputs or conversely recognize when they are wrong \cite{finale2017towards}.

Ensembles of decision trees such as random forests \cite{breiman2001random} and boosted trees  \cite{friedman2001greedy} 
perform well across a variety of problems \cite{caruana2006empirical}. However, while their decision tree components may be interpretable \cite{freitas2014comprehensible}, this is no longer true for ensembles with hundreds or thousands of trees. Current attempts to interpret tree ensembles include seeking one tree that best represents the ensemble \cite{hara2016making,zhou2018approximation}, model-agnostic explanations not exclusive to tree ensembles \cite{lime}, feature importance \cite{ishwaran2007variable}, partial dependence plots \cite{friedman2001greedy}, etc. However, many of these describe how features affect predictions, and their complexity increases with the number of features.

\begin{figure*}[ht!]
    \centering
    \vspace{-0.35cm}
    \includegraphics[width=0.75\textwidth]{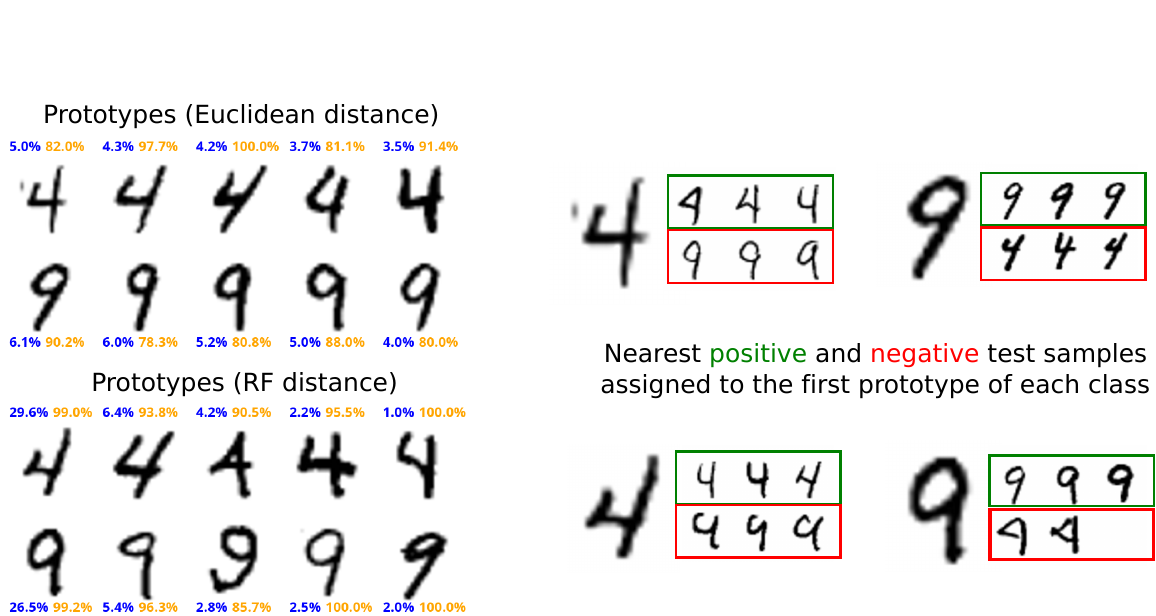}
    \caption{\emph{Left:} Prototypes with largest coverage for the classes 4 and 9 on the MNIST dataset when using Euclidean distance and random forest distance to find the prototypes. The number in blue denotes the coverage of the prototypes (percentage of test points assigned to that prototype) while the number in orange denotes the accuracy of the prototype (percentage of points assigned to that prototype that have the same label as the prototype). 
    Note the hooked 9 and closed 4, which are only captured by RF distance but not Euclidean distance.
    \emph{Right:} Nearest correct and incorrect test points assigned to the top prototypes. The first row denotes points with the same label as the prototype, second row are points incorrectly classified by the prototype.}
    \label{fig:fig1}
\end{figure*}

Prototypes are representative points that provide a condensed view of a dataset \cite{hart1968condensed,bien2011prototype}. The value of prototypes for case-based reasoning \cite{richter2005case} has been discussed in studies of human
decision making \cite{kim2014bayesian}. Prototypes have also been used to summarize large datasets  \cite{mirzasoleiman2013distributed} when not all points can be inspected. In this paper, we propose an alternative to feature-based explanations for tree ensemble classifiers: rather than explaining which features led to a certain class being predicted, we propose to explain a prediction by presenting similar points that ``represent'' that class (Figure \ref{fig:fig1}). Since these prototypes will be identified using distance functions derived from the tree ensemble, we call them tree space prototypes.

A key question is how to define similarity. Unsupervised distances such as Euclidean distance in feature space do not capture relationships between features and labels, whether actual or predicted. Instead, we need a distance that takes into account: (1) the predictions made by the tree ensemble; (2) how the predictions came about (i.e. how the tree ensemble used the features to arrive at the predictions). Such a distance has been defined for random forest models (RF) where each tree contributes equally to the overall prediction. We generalize this to gradient boosted trees (GBTs), where individual trees that make up a GBT model can have different contributions to the overall prediction. 

By adapting a known approximation algorithm for the $k$-medoids problem, we can efficiently search for prototypes that are 
``central'' for a class according to these proximity functions. Nearest-prototype classifiers using these prototypes sometimes even exceed the accuracy of the original tree ensemble. However, this algorithm has no notion of when one class may benefit from more prototypes than another class (e.g. if one class is more complex: consider for instance a disease which affects many different types of individuals, but does not affect one type of individual, so the class of sick individuals is more complex to characterize than the class of healthy individuals). Hence, we introduce new \emph{class-aware} prototype selection methods with the flexibility to choose a variable number of prototypes per class. 

To evaluate the interpretability of tree space prototypes, we conducted a user study in which human subjects anticipated the output of a tree ensemble classifier on a dataset of car fuel efficiency, using either prototypes or Shapley values, a popular feature attribution method \cite{Lundberg2017unified}. Our results suggest ($p\approx 0.035$) that prototypes convey better understanding of the tree ensemble classifier's behavior.

To summarize, the contributions of this paper are: (1) An alternative approach to interpreting tree ensemble classifiers by selecting 
representative points -- prototypes -- for each class; (2) a new distance function for GBT models; (3) new prototype selection methods with theoretical guarantees, that have the flexibility to choose a different number of prototypes in each class.

\section{Background and Notation}
\subsection{RF Distance} 
Let $t$ be the number of trees in the RF model.
The $i$th tree ($i\in [t]$) has $\tau_i$ leaves, each of which represents a region $R_{j,i}$ ($j\in [\tau_i]$) of feature space. Each individual tree induces a classifier
$ c^{\text{Tree}}_i(s) = \sum_{j=1}^{\tau_i} \alpha_{j,i} \mathbb{I}(s\in R_{j,i}), $
where $\alpha_{j,i}$ is the predicted value in the $j$th leaf of the $i$th tree (for binary classification, this is just the proportion of points in that leaf with label 1) and $\mathbb{I}$ is the indicator function. The RF classifier is
the average of this, taken over all trees:
$$ c^{\text{RF}}(s) = \frac1t \sum_{i=1}^t c^{\text{Tree}}_i. $$

 Using the above notation, we can re-write the random forest classifier as
 \[
 c^{RF}(s) = \frac{1}{t} \sum_{i=1}^t \sum_{j =1}^{\tau_i} \frac{1}{N} \sum_{k=1}^N \mathbb{I}(s \in R_{j,i}) \mathbb{I}(s_k \in R_{j,i}) y_k = \frac{1}{N} \sum_{k=1}^N K(s,s_k) y_k
 \]
 for the kernel
 \[
 K(s,s') = \frac{1}{t} \sum_{i=1}^t \sum_{j =1}^{\tau_i}   \mathbb{I}(s \in R_{j,i}) \mathbb{I}(s' \in R_{j,i}).
 \]
This connection has allowed the study of random forests in which tree structure is generated independently of the data; in particular \cite{scornet2016random} provides explicit formulae for the corresponding $K(s,s')$, although this is much more challenging for supervised trees which adapt to the contours of the underlying response. This same representation results in the proximity function between two points:
 
\begin{definition}{\cite{breiman2002rf}}
\label{def:rfprox}
The RF proximity of a pair of points is an unweighted average of the number of trees in the RF model in which the points end up in the same leaf:x
\begin{eqnarray}& &\mathsf{proximity}^\mathrm{RF}(s,s') \nonumber \\ &=& \frac{1}{t} \displaystyle \sum_{i=1}^t \displaystyle \sum_{j=1}^{\tau_i} \mathbb{I}(s \in R_{j,i})\mathbb{I}(s' \in R_{j,i}). \label{eqn:rfprox} \end{eqnarray}

The RF distance between a pair of points is then: 
$$d^{\mathrm{RF}}(s,s') = 1 - \mathsf{proximity}^{\mathrm{RF}}(s,s').$$
\end{definition}

Since the regions $\{ R_{j,i} \}_{j=1}^{\tau_i}$ partition the feature space, each point $s\in S$ can be in at most one region, and so the inner sum in Equation \eqref{eqn:rfprox} takes on value $0$ or $1$ for each tree. Thus the proximity, as a convex combination of these, lies between 0 and 1, and so does the distance function. It is easily confirmed that the proximity of a point to itself is 1, and hence $d(s,s)=0$, but it should be noted that $d$ is not in general a metric, but a pseudosemimetric as it does not satisfy the triangle inequality -- as noted by \cite{xiong2012random}. This it not uncommon in the metric learning literature, and in fact, no locally adaptive distance -- distance that varies across feature space \cite{lin2006random} -- can satisfy the triangle inequality \cite{xiong2012random}. Later, we will adapt RF distance to construct a distance function for GBTs.

\subsection{The $k$-Medoids Problem}
Given a proximity function, it is natural to construct a classifier by taking those points that are particularly close to some point or region considered representative (prototypical) of a class to belong to that class. How should these prototypes be selected? 

For accuracy, we would want every point in a class to be closer to a prototype of that class than any prototype of another class. This is generally hard; a more tractable related approach is to instead seek to simply make the points in each class as close to a prototype as possible. If we further take the tradeoff between different points' distance from a prototype to be linear, this is known as the \emph{$k$-medoids clustering problem} (see e.g. \cite{bien2011prototype} for another application to prototypes). 
The objective of this problem is to find a subset $M\subseteq S$ of medoids, $|M|=k$,
such that the sum distance from each object to the nearest medoid is minimized. Formally, we seek to minimize the objective function
\begin{equation} f(M) = \sum_{s\in S} \min_{m\in M} d(s,m).
\label{eqn:noclassobjective}
\end{equation}
This problem is known to be NP-hard \cite{papad1981}. However, \cite{gomeskrause} present a greedy algorithm that starts with an empty set and repeatedly adds the single point $s\in S\setminus M$ that increases the value of a related function by the most, which they show produces a reasonable approximation in polynomial time.

If the points are labelled by a classifier, it is natural to only consider, for each point, medoids that belong to the same class. Thus, we
define the \emph{$q$-classwise $k$-medoids problem} as finding the subset $M\subseteq S$ of $k$ medoids
such that the sum distance from each point to the nearest medoid \emph{of the same class} is minimized, i.e. that minimizes
\begin{equation} f(M) = \sum_{s\in S} \min_{m\in M: c(m)=c(s)} d(s,m).
\label{eqn:objective}
\end{equation}

Even in the presence of multiple classes, it is  possible to use the single-class algorithm of \cite{gomeskrause} by applying it separately to every class in turn to generate $k_1$, $\ldots$, $k_q$ prototypes for each class ($\sum_i k_i = k$). However, it is not clear what the right choice of $k_i$ for each class is, and one could easily lose accuracy by overprovisioning one compact class that would be adequately covered by a small number of prototypes while not having sufficiently many prototypes for another class whose points are spread into many clusters. With the naive choice that $k_1=\ldots=k_q=k/q$, we call this the \textcolor{blue}{uniform greedy submodular (SM-U)} prototype selection method, and use it as one of our baselines.

However, it turns out that an analysis similar to that for the single-class case can also be applied directly to the $q$-classwise objective function. Based on this, we will introduce a greedy algorithm that operates on all classes in the $q$-classwise $k$-medoids problem simultaneously. Since this algorithm in effect chooses the class where adding another prototype yields the largest improvement, we will call it \textit{adaptive} in contrast with the uniform algorithm.

\section{Method}

Our goal is to find prototypes for tree ensemble classifiers. In this section, we describe two methodological contributions of this paper: defining a distance function for GBT models, and new, adaptive prototype selection methods that choose a variable number of prototypes based on which class could benefit the most from another prototype.

\subsection{Constructing a Distance Function for GBT}

We start by considering the prediction function of the GBT classifier, which is learned iteratively: 
 $$ c^{\text{GBT}}_i(s) = c^{\text{GBT}}_{i-1}(s) + \gamma_i c^{\text{Tree}}_i(s) $$
 where the initial value $c^{\text{GBT}}_0$ is initialized, depending on implementation, as zero, or the fraction of elements of $S$ with label 1 in the case of binary classification, etc. $\gamma_i$ is a step size, typically found using line-search, that provides a correction to account for the quadratic approximation to the loss that is used by gradient boosting. The GBT classifier then is the one that incorporates all $t$ trees:
 $$ c^{\text{GBT}}(s) = c^{\text{GBT}}_t(s). $$
 
Unlike RF, GBTs cannot be expressed directly as kernel methods: the values in each leaf are not given by averages of the corresponding responses. Further, each tree is no longer generated by an identical process or contributes equally to the prediction. Hence, each tree can no longer be weighted equally, unlike in RF models.
Instead, we propose that a natural way is to weigh the contribution of each tree to the proximity function by the size of its contribution to the overall prediction. By using the $L_2$ norm to measure size, we arrive at the following definition:

\begin{definition} The GBT proximity of a pair of points is a weighted average of the number of trees in the GBT model in which the points end up in the same leaf:
\begin{eqnarray*}
 & & \mathsf{proximity}^{\mathrm{GBT}}(s,s') \\
  &=& \sum_{i=1}^t \sum_{j=1}^{\tau_i} \frac{w_i}{\sum_{i=1}^t w_i} \mathbb{I}(s \in R_{j,i}) \mathbb{I}(s' \in R_{j,i}),
  \end{eqnarray*}
where the $i$th tree's weight $w_i$ is $$w_i = \gamma_i^2 \cdot \mathrm{Var}\{c_i^{\mathrm{Tree}}(s): s \in S\}.$$

The GBT distance between a pair of points is then: 
$$ d^{\mathrm{GBT}}(s,s') = 1 - \mathsf{proximity}^{\mathrm{GBT}}(s,s'). $$
\end{definition}

The choice of the $L_2$ norm to measure the size of a tree's contribution to the overall prediction has a natural equivalence to measuring the variance among the predictions made by $c_i^{\text{Tree}}(s)$. 
As an alternative to the $L^2$ norm used here, one may instead consider the $L^1$ norm, which we leave for future work. In Section \ref{sec:understand_gbt_distance}, we study the implications of selecting this weight on the constructed distance.

\subsection{Adaptive Prototype Selection Methods}
We now introduce two new prototype selection methods that exploit approximation guarantees for submodular objective functions, and one that tries to directly optimize for accuracy.

Our goal is to find a good approximately optimal solution for the $q$-classwise $k$-medoids problem (\ref{eqn:objective}). We will achieve this by using a greedy algorithm on an appropriate non-negative, monotone, submodular function (see Prop. \ref{prop:submodular}). However, the function (\ref{eqn:objective}) itself is not monotone submodular: in fact, adding more prototypes to $M$ \emph{decreases} the value of $f(M)$. This can be averted by negating $f$, but then the function will take non-positive values. Therefore, adapting an idea of \cite{gomeskrause}, we will define a related function $g$ as
\begin{equation}
    g(M) = f(P)-f(P\cup M), \label{eqn:objectivefixed}
\end{equation}
where $P$ is a set of \emph{phantom exemplars}, one from each class. In order to get the best possible theoretical guarantee on the approximation (Section \ref{sec:analysis}), this set needs to be chosen in a particular fashion.
The resulting algorithm is Algorithm \ref{alg:submodularall}.

\begin{algorithm}
\caption{\textcolor{greenmatplotlib}{Adaptive greedy submodular prototype selection (SM-A)}}
\label{alg:submodularall}
\KwIn{Set of points $S$, distance function $d:S^2\rightarrow [0,1]$, class assignment $c:S\rightarrow[q]$}
\KwOut{Set of prototypes $M$, $|M|=k$}
\nl Create set of phantom exemplars $P=\{p_1,\ldots,p_q\}$ and set $d(p_i,s)=d(s,p_i)=1$ for all $s$\\
\nl $M\leftarrow \varnothing$\\
\nl \For{i=1 \KwTo k}{
\nl $s^* \leftarrow \underset{s\in S} {\arg\max} \left[ f(P)-f(P\cup M\cup \{s\}) \right]$\\
\nl $M\leftarrow M\cup \{s^*\}$
}
\end{algorithm}

We also consider a variant of this algorithm that we call \textcolor{purple}{weighted adaptive greedy submodular (SM-WA)}, in which each class is
weighed differently: line 4 is replaced by
$$ s^* \leftarrow \underset{s\in S} {\arg\max} \frac{1}{|C(s)|}\left[ f(P)-f(P\cup M\cup \{s\}) \right], $$
where $C(S)$ denotes all points in $S$ that are in the same class as $s$.
It is easily verified that this objective function is also submodular.

\subsection{Supervised Greedy Prototype Selection}

Instead of optimizing the $k$-medoids value function $f$ of equation (\ref{eqn:objective}), we can instead directly pick prototypes, in a greedy fashion, that yield the best (training or validation set) improvement in classification performance. The resulting method, which we call \textcolor{red}{supervised greedy (SG)}, beats the unsupervised $k$-medoids approaches in terms of accuracy in several cases (Table \ref{tab:validated_prototype}), but we do not know of any theoretical guarantees that it satisfies, as these accuracy metrics are not submodular. This is nearly identical to Algorithm \ref{alg:submodularall}, except that line 1 is unnecessary and we replace line 4 with $$s^* \leftarrow \underset{s\in S} {\arg\max} \left[ \mathsf{accuracy}(S,M\cup \{s\}) \right],$$
where $\mathsf{accuracy}$ denotes the accuracy metric used for evaluation.

\section{Theoretical Analysis}
\label{sec:analysis}

We will now briefly review the rationale behind the design of Algorithm \ref{alg:submodularall} and derive an approximation guarantee for it.
Optimization problems such as (\ref{eqn:objective}) are often approached using approximation algorithms that are guaranteed to find solutions within some factor of the optimum. 
Previous work on $k$-medoids  \cite{gomeskrause,mirzasoleiman2013distributed} has achieved this by identifying a related positive monotone submodular function and finding a good element of its domain by greedy search. Such an element is guaranteed to be within a factor of $(1-1/e)$ of the optimum for that function, where $e$ is the Euler constant. We quickly review the relevant result.

\begin{definition} A function $f:\mathcal{P}(S) \rightarrow \mathbb{R}$ that maps subsets of $S$ to reals is \emph{monotone}
 if $f(X)\leq f(Y)$ whenever $X\subseteq Y$. It is \emph{submodular} if whenever $X\subseteq Y$, adding
 a particular element $s\in S$ to $Y$ is not more useful than adding it to $X$:
 $$f(Y\cup \{s\})-f(Y) \leq f(X\cup \{s\})-f(X). $$  
 \label{def:submod}
 \end{definition}
 \begin{proposition} \label{prop:submodular}
 {\cite{nemhauser1978analysis}} Suppose $f:\mathcal{P}(S)\rightarrow \mathbb{R}^+$ is a non-negative monotone submodular function.
 Let $T_0=\varnothing$ and $$T_{i}=T_{i-1} \cup \underset{s\in S}{\arg\max} f(T_{i-1}\cup \{s\})$$ be the result
 of greedily maximizing $f$ for $i$ steps. Also, let $$T_i^* = \underset{T\subset S: |T|=k}{\arg\max} f(T)$$ be 
 the set of size $i$ that maximizes $f$.
 Then $$f(T_i) \geq (1-1/e) f(T_i^*).$$ 
 \end{proposition}

We want to derive a similar approximation guarantee for Algorithm \ref{alg:submodularall}. To that end, we first need to show that $g$ satisfies the necessary conditions.

 \begin{lemma} The objective function (\ref{eqn:objectivefixed}) is non-negative, monotone and submodular.
 \end{lemma}
 \begin{proof} See appendix. 
 \end{proof}
 
 By selecting the set of phantom exemplars $P$ in such a fashion that $d(p,s)\geq d(s',s)$ for all $p\in P$ and $s,s'\in S$, we ensure that $f(T\cup P)=f(T)$ for all nonempty sets $T\subseteq S$. Hence, the set $T_i^*$ that maximizes $g$ among all sets of size $i$ also minimizes $f$ among all such sets.
 
  Let $T_i$ be the result of running the greedy maximization algorithm on (\ref{eqn:objectivefixed}) for $i$ steps, and $f$ be the original objective function (\ref{eqn:objective}). Then by Prop. \ref{prop:submodular} and choice of $P$,
 $$ f(T_i) \leq f(P) + (1-1/e) (f(T_i^*) - f(P) ), $$
 i.e. the approximation $T_i$ takes us $1-1/e$ of the way from $f(P)$ to the optimum. 
 Crucially, this means that the approximation guarantee depends on $f(P)$, i.e. how good the phantom exemplars alone would be as a solution to the $q$-classwise $k$-medoids problem.
 
 \textbf{Complexity analysis.}
 Evaluating $f(M)$ takes time $O(|S||M|\cdot T(d))$, where $T(d)$ is the time to compute the distance $d(s,m)$ for a single pair of points. This computation can be made efficient by prepopulating an $|S|\times|S|$ matrix with all pairwise distances, and then simply implementing $d$ as an array lookup.
 The same complexity bound applies to calculating $\mathsf{accuracy}(S,M)$, which iterates over $|S|$ points and finds the $d$-closest of $|M|$ medoids to check if it belongs to the correct class.
 The submodular (\textcolor{greenmatplotlib}{SM-A}, \textcolor{purple}{SM-WA}) and supervised greedy (\textcolor{red}{SG}) variants of Algorithm \ref{alg:submodularall} essentially only differ in whether they invoke $f$ or $\mathsf{accuracy}$ with an $O(k)$-sized set $M$ of medoids in the $\arg\max$ (Algorithm \ref{alg:submodularall}, Line 4). Either way, this $\arg\max$ is over $|S|$ points, and the loop runs for $k$ iterations. Therefore, all our instantiations of Algorithm 1 have time complexity $O(|S|^2 k^2 \cdot T(d))$.

\section{Related Work}
\textbf{Tree ensemble distance.} 
Breiman and Cutler defined RF proximity in the documentation accompanying their software \cite{breiman2002rf}. It is common to set distance as $1 - \mathrm{proximity}$ \cite{zhao2016propensity,xiong2012random,shi2006unsupervised}, as we do in this paper. RF proximity has found a variety of applications, including clustering \cite{shi2006unsupervised}, outlier detection \cite{zhou2015two}, imputation \cite{stekhoven2015missforest}, etc., however less is known of its theoretical properties. The connection between random forests and kernel methods has been pointed out \cite{lin2006random,scornet2016random} and proximity itself can be expressed as a kernel \cite{louppe2014understanding}. While we were inspired by RF distance, to the best of our knowledge, our paper presents the first proposal for GBT distance and the first method to seek prototypes for GBT models. 

Not many RF implementations provide prototypes. The exceptions are the R \texttt{randomForest} package \cite{liaw2002classification} and RAFT, a random forests visualization tool \cite{breiman2002rf}. The RAFT documentation describes a heuristic prototype-finding procedure that is partially implemented in the \texttt{randomForest} package. It generates a single new point not from the dataset, a distinct goal from ours, which is to select a subset of representative points from existing points.

\textbf{Prototype selection.} 
There is a long line of literature on prototype selection methods, also known as instance reduction, data summarization, exemplar extraction, etc. We point the reader to the review by \cite{garcia2011prototype} that suggested that prototype selection methods can be grouped into three
categories: condensation \cite{hart1968condensed},
edition \cite{wilson1972}, 
or hybrid methods that remove both noisy and redundant points from the prototype selection set. We briefly mention a few methods: $k$-medoids clustering is a classic problem for which different algorithms have been proposed, such as PAM \cite{kaufman1987clustering} and greedy submodular approaches \cite{lin2011class,gomeskrause}, which we compare against and extend by adding the flexibility to choose varying numbers of prototypes by class. Kim et al. used a similar greedy submodular approach with maximum mean discrepancy objective to select prototypes and criticisms \cite{kim2016examples}; we provide a comparison to their prototype (not criticisms) selection method. We do not compare against set cover methods \cite{bien2011prototype} as they tend to select significantly more prototypes than $k$-medoids \cite{bien2011prototype} to achieve their objective of maximal coverage, at the cost of interpretability. 

\textbf{Point-based explanations.} Besides feature-based explanations such as Shapley values \cite{Lundberg2017unified} and LIME \cite{lime}, point-based explanations have been proposed to explain model predictions. Examples include counterfactual explanations that determine the changes necessary to flip a point's prediction \cite{wachter2017counterfactual}, models that automatically provide prototypes \cite{kim2014bayesian,li2018deep}, and identifying points most ``influential'' for a prediction \cite{koh2017understanding,yeh2018representer,khanna2018interpreting}. There is a subtle distinction between prototype selection methods and influential point methods, as points that best represent a class (prototypes) may not be the most influential. Moreover, since trees are not differentiable except trivially within each node, influential point methods that typically take gradients of loss functions are not easily applicable. Hence we do not compare against them, but mention them for completeness. 

The work most similar in spirit to ours is by Caruana et al. \cite{caruana1999case} who proposed to generate case-based explanations for non-case-based learning models such as neural networks and decision trees. However, unlike this paper, they do not take advantage of naturally-learned distance functions from these models.

\begin{table*}
\centering
\footnotesize
\begin{tabular}{lllllllll}
\toprule
Model &  \multicolumn{2}{l}{Prototype Selection Method} & Breastcancer & Diabetes & T-COMPAS & RHC & MNIST 4-9 & CALTECH256 G-M\\
\midrule
\multirow{8}{*}{RF} & \multicolumn{2}{l}{None (original tree ensemble)} & 0.92 & 0.72 & 0.56 & 0.68 & 0.97 & 0.81\\
\cdashlinelr{2-9}
 & \multirow{3}{*}{Baselines} &\textcolor{orange}{1-NN} & 0.91 (341) & 0.67 (460) & 0.57 (600) & 0.65 (3441) & \textbf{0.97} (3000) & 0.78 (129)\\
 & & \textcolor{blue}{SM-U} & \textbf{0.92} (12) & 0.76 (5) & 0.61 (10) & 0.69 (11) & \textbf{0.97} (187) & 0.83 (16)\\
 & & \textcolor{olive}{Kim et al \cite{kim2016examples}} & \textbf{0.92} (19) & 0.68 (36) & 0.62 (32) & 0.68 (3) & 0.96 (65) & 0.81 (6)\\
\cdashlinelr{2-9}
 & \multirow{3}{*}{Proposed} & \textcolor{red}{SG} & 0.90 (4) & \textbf{0.77} (5) & \textbf{0.68} (5) & 0.66 (12) & \textbf{0.97} (18) & 0.83 (3)\\
 & & \textcolor{green}{SM-A} & \textbf{0.92} (11) & \textbf{0.77} (4) & 0.57 (15) & 0.68 (9) & \textbf{0.97} (163) & 0.79 (15)\\
 & & \textcolor{purple}{SM-WA} & \textbf{0.92} (15) & \textbf{0.77} (6) & 0.60 (13) & \textbf{0.69} (13) & \textbf{0.97} (243) & \textbf{0.86} (18)\\
\midrule
\multirow{8}{*}{GBT} & \multicolumn{2}{l}{None (original tree ensemble)} & 0.94 & 0.69 & 0.55 & 0.70 & 0.97 & 0.84\\
\cdashlinelr{2-9}
 & \multirow{3}{*}{Baselines} &\textcolor{orange}{1-NN} & 0.92 (341) & 0.70 (460) & 0.58 (600) & 0.65 (3441) & \textbf{0.97} (3000) & 0.84 (129)\\
 & & \textcolor{blue}{SM-U} & 0.92 (21) & 0.70 (12) & 0.53 (26) & 0.65 (62) & 0.96 (249) & 0.84 (11)\\
 & & \textcolor{olive}{Kim et al \cite{kim2016examples}} & 0.94 (6) & 0.66 (16) & 0.63 (53) & 0.61 (33) & 0.94 (45) & \textbf{0.86} (26)\\
\cdashlinelr{2-9}
 & \multirow{3}{*}{Proposed} & \textcolor{red}{SG} & \textbf{0.95} (3) & \textbf{0.78} (4) & \textbf{0.67} (5) & \textbf{0.69} (15) & 0.96 (23) & 0.82 (2)\\
 & & \textcolor{green}{SM-A} & 0.92 (22) & 0.69 (12) & 0.57 (4) & 0.65 (4) & 0.96 (247) & 0.84 (11)\\
 & & \textcolor{purple}{SM-WA} & 0.92 (20) & 0.69 (12) & 0.56 (27) & 0.65 (4) & 0.96 (261) & 0.84 (11)\\
\midrule
\multirow{7}{*}{EUCL} & \multirow{3}{*}{Baselines} & \textcolor{orange}{1-NN} & \textbf{0.91} (341) & 0.68 (460) & 0.53 (600) & 0.59 (3441) & \textbf{0.96} (3000) & 0.81 (129)\\
 & & \textcolor{blue}{SM-U} & 0.89 (19) & 0.71 (26) & 0.51 (62) & 0.61 (40) & 0.93 (313) & 0.82 (6)\\
 & & \textcolor{olive}{Kim et al \cite{kim2016examples}} & 0.88 (60) & 0.66 (29) & 0.51 (49) & 0.60 (5) & 0.92 (384) & 0.78 (64)\\
\cdashlinelr{2-9}
 & \multirow{3}{*}{Proposed} & \textcolor{red}{SG} & 0.87 (3) & \textbf{0.75} (4) & \textbf{0.58} (22) & \textbf{0.67} (19) & 0.90 (65) & 0.74 (9)\\
 & & \textcolor{green}{SM-A} & 0.88 (18) & 0.68 (3) & 0.52 (51) & 0.60 (8) & 0.93 (377) & \textbf{0.88} (6)\\
 & & \textcolor{purple}{SM-WA} & \textbf{0.91} (15) & 0.73 (5) & 0.51 (61) & 0.61 (33) & 0.93 (380) & \textbf{0.88} (6)\\
\bottomrule
\end{tabular}
\caption{Best test-set balanced accuracy with corresponding optimal number of prototypes, $k$, in parentheses. Three distances are provided: Random Forest (RF), Gradient Boosted Tree (GBT), and Euclidean (EUCL). We compare the proposed \textcolor{red}{supervised greedy (SG)}, \textcolor{greenmatplotlib}{adaptive greedy submodular (SM-A)}, and \textcolor{violet}{weighted adaptive greedy submodular (SM-WA)} prototype selection methods against the \textcolor{blue}{uniform greedy submodular (SM-U)} and \textcolor{orange}{1-NN} baselines. We also compare our results with the greedy prototype selection method from \textcolor{olive}{Kim et al \cite{kim2016examples}}. Best results for each dataset and distance in \textbf{bold}.}
\label{tab:validated_prototype}
\end{table*}

\section{Experimental Results}
We evaluate the proposed prototype selection methods and tree ensemble distances quantitatively as well as qualitatively. We also describe results from a user study that demonstrates that humans are able to use prototypes effectively.

\textbf{Datasets.} We use multiple image and tabular datasets with binary classification labels. For image datasets, we select two standard image classification benchmarks, MNIST  and CALTECH-256. The goal in MNIST is to recognize handwritten digits \cite{mnist}; the goal in CALTECH-256 is to predict one of 256 object categories for an image \cite{caltech256}. For both datasets, we select two classes that are either easily-confused \cite{hadsell2006dimensionality} or  visually-similar -- digits 4 and 9 in MNIST, guitar and mandolin in CALTECH-256 -- to evaluate our prototypes on not just easily predicted classes, but also classes commonly confused by the model. For MNIST, we use the raw pixel values as features. For CALTECH-256, we extracted deep features using a ResNet-50 model pre-trained on ImageNet. 

For tabular datasets, we selected four datasets from critical domains such as healthcare and criminal justice where the need for interpretability has been suggested. These four datasets are: \texttt{sklearn} breastcancer and UCI diabetes, where the prediction task is to predict incidence of that disease, the Right Heart Catherization (RHC) dataset (\url{http://biostat.mc.vanderbilt.edu/wiki/pub/Main/DataSets/rhc.html}), on the impact of performing a medical procedure on patients, and T-COMPAS \cite{dressel2018accuracy}, a dataset that examines if COMPAS risk scores  \cite{propublica2018compas} agree with Mechanical Turk workers' predictions of recidivism \cite{dressel2018accuracy}. The first two datasets are common tabular data benchmarks, and we selected the last two datasets because their prediction tasks are known to be hard \cite{dressel2018accuracy}, again to validate our prototypes on a diversity of cases, not just easy ones.

\textbf{Metrics.} Since some datasets are imbalanced, we use balanced accuracy \cite{balancedaccuracy} as our primary performance metric.
We do not use ranking metrics such as AUC because nearest-neighbor classifiers do not output scores. We also count the number of prototypes selected.

\textbf{Training and tuning tree ensembles.} Whenever a fixed train-test split was not provided (i.e. for all datasets besides MNIST, where the training set has 60,000 images and the test set has 10,000 images), we created 60-20-20\% training-validation-test splits. For RF, we use Python's \texttt{scikit-learn} package, training random forests with 1,000 trees without restricting maximum tree depth. We cross-validated the number of features to consider when looking for the best split ($\sqrt{p}$, $0.33p$, $0.5p$, $0.7p$, where $p$ is the number of features, and the constant $7$). For GBT, we modified \texttt{scikit-learn} to train GBT models with one gamma multiplier per tree. 
We cross-validated the number of trees (up to 200), maximum tree depth (3 to 5), and learning rate (0.1 or 0.01).

\textbf{Implementations.} For the comparison to Kim et al. \cite{kim2016examples}, we use the authors' code for greedy prototypes (not criticisms) selection  provided at \url{https://github.com/BeenKim/MMD-critic}. To obtain Shapley values \cite{Lundberg2017unified} and associated graphs for the user study, we use the authors' Python Shap package which can be found at \url{https://github.com/slundberg/shap}. 

\subsection{Quantitative Evaluation}
We quantitatively evaluate the selected prototypes by using them in a \emph{nearest-prototype classifier} \cite{bien2011prototype,kim2016examples}. This is in line with recent ideas on evaluating explanations by checking their accuracy on independent test-data \cite{lime}. 

In general, the accuracy of each prototype selection method varies non-monotonically with $k$, the number of prototypes, suggesting that $k$ should be tuned. Moreover, different selection methods operate at different regimes, e.g, \textcolor{red}{supervised greedy (SG)} obtains very good results with very few prototypes, but is sometimes outperformed by other methods when they use a larger number of prototypes.
Comparing different selection methods using the same $k$ may therefore not accurately characterize each method. Instead, we follow \cite{bien2011prototype}, tuning $k$ separately for each prototype selection method and dataset, and comparing different methods at their optimal $k$. 

It should be noted that in the limit, when $k$ equals the size of the training set, any nearest-prototype classifier simply reduces to the \textcolor{orange}{1-nearest-neighbors (1-NN)} classifier. This classifier is hence one of our s, along with the original tree ensembles and the \textcolor{blue}{uniform greedy submodular (SM-U)} approach. We also compare RF and GBT distance functions to Euclidean distance in feature space. 

Table \ref{tab:validated_prototype} summarizes the nearest-prototype classifier results. It provides test set balanced accuracy at the optimal number of prototypes, $k$, tuned separately for each prototype selection method. We make several observations:
\begin{enumerate}[(1)]
\item For all datasets
besides MNIST 4-9,
at least one prototype selection method outperformed 1-NN, suggesting the value of prototype selection not just for data condensation and interpretability (reducing the number of points that need to be shown to a user), but also classification accuracy. 
\item Tree ensemble distances, as supervised distances, tend to outperform Euclidean distance. 
\item \textcolor{purple}{SM-WA} is competitive against \textcolor{blue}{SM-U}.
\item Despite the lack of theoretical guarantees and simplicity of the method, \textcolor{red}{SG} had clear advantages on a number of datasets with high achieving high accuracy.
\item \textcolor{red}{SG} tends to select fewer prototypes than the other methods.
\item At least one of our proposed prototype selection methods outperforms \textcolor{olive}{Kim et al. \cite{kim2016examples}}, with higher accuracy, lower prototype count, or both on each dataset.
\end{enumerate}

\begin{figure*}
\vspace{-0.5cm}
    \centering
    \includegraphics[width=0.4\textwidth]{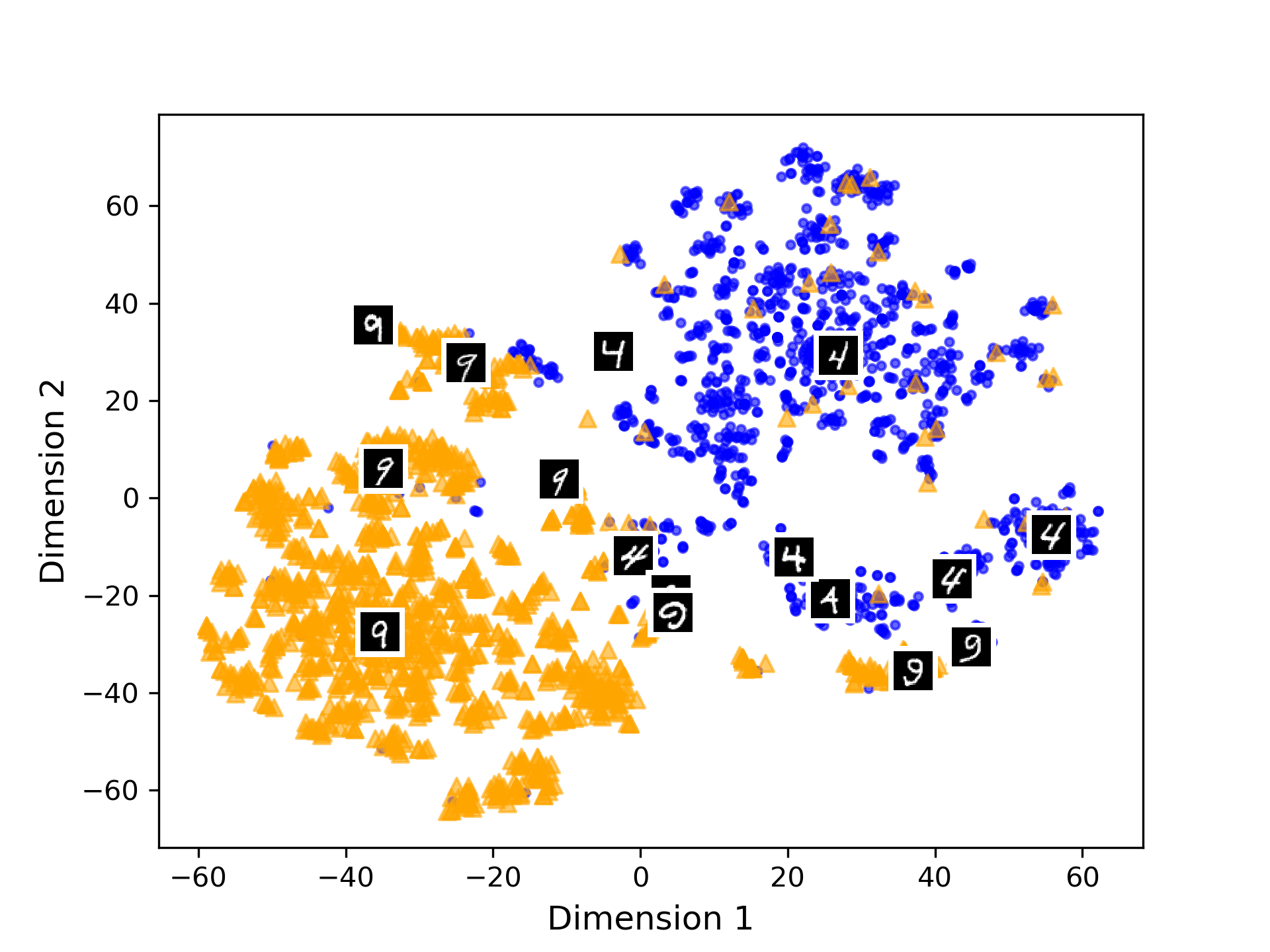}
    \includegraphics[width=0.4\textwidth]{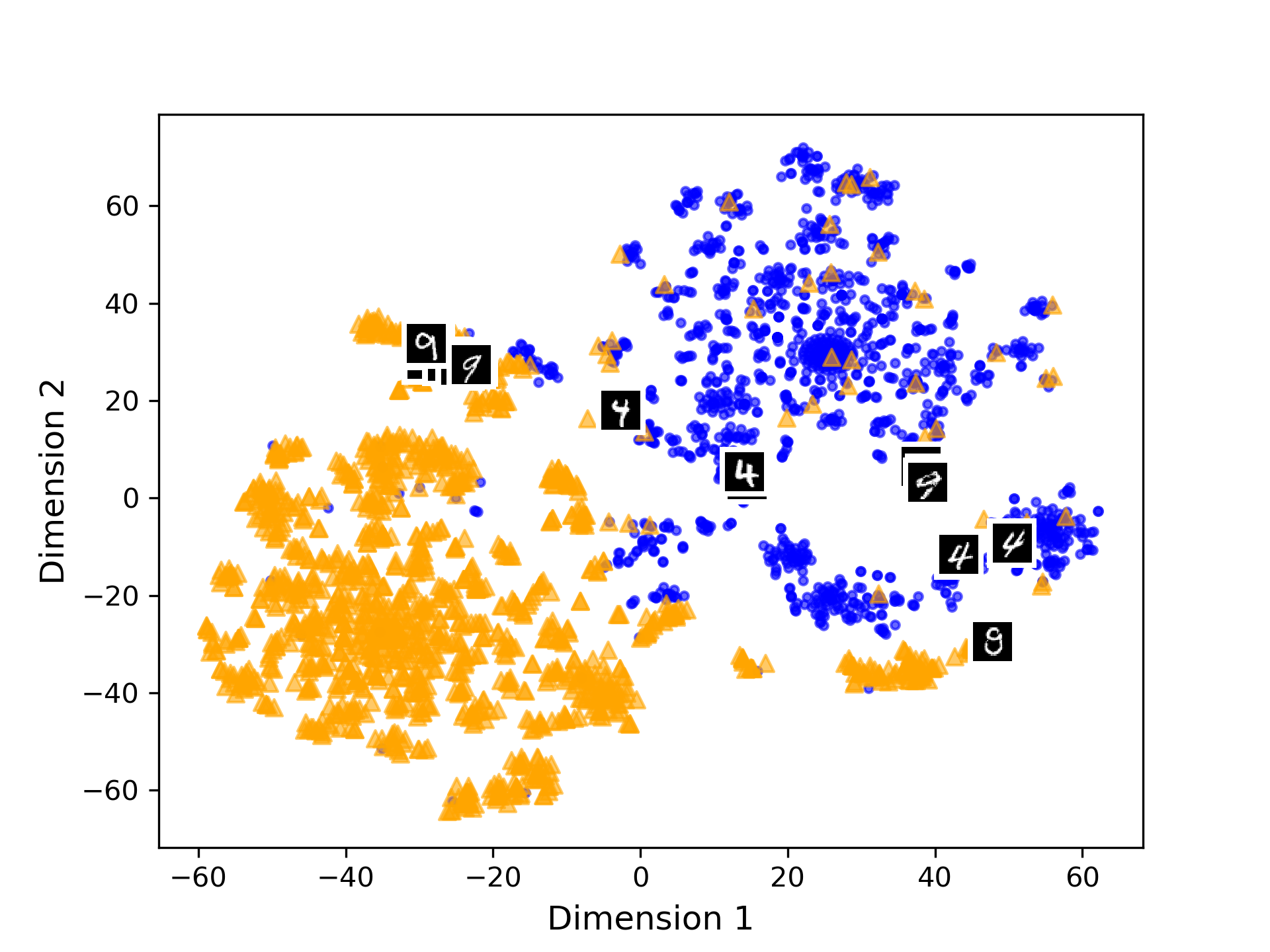}
     \caption{Visualization of distances using t-sne 
for optimal RF with mean depth 16 on the MNIST 4-9 dataset, using the \textcolor{greenmatplotlib}{adaptive greedy submodular (SM-A)} prototype selection method (left) and \textcolor{red}{supervised greedy (SG)} (right) method. \textcolor{orange}{Orange} represents the digit 9, \textcolor{blue}{blue} represents the digit 4. and the black and white images are prototypes.}
    \label{fig:comparingoptimizersmnist}
\end{figure*}

\subsection{Visualizing Distances and Prototypes}
Figure \ref{fig:comparingoptimizersmnist} visualizes RF distance for the MNIST 4-9 dataset embedded in a two-dimensional space using t-sne \cite{maaten2008visualizing} . On the left side are prototypes found by \textcolor{greenmatplotlib}{SM-A}; on the right side are prototypes selected by \textcolor{red}{SG}. 

We see that the prototypes selected by \textcolor{greenmatplotlib}{SM-A} cover the space of points well, which is not the case for \textcolor{red}{SG}. We confirm this by computing the distance from each point to its nearest-prototype (\textcolor{greenmatplotlib}{SM-A}: mean 0.23, sd 0.26; \textcolor{red}{SG}: mean 0.61, sd 0.23). Instead, the prototypes selected by \textcolor{red}{SG} are on the border between the two classes, and it is common to see these prototypes alternating (i.e. a 9 prototype followed by a nearby point of class 4 being selected as a prototype). 

This has an intuitive explanation: as the only supervised prototype selection method, to maximize accuracy and minimize the chances of misclassification, \textcolor{red}{SG} selects discriminative prototypes that can separate points that are of different classes yet are close to each other, while the other proposed prototype selection methods, being non-supervised, do not select prototypes discriminatively. 

\subsection{Understanding GBT Distance}
\label{sec:understand_gbt_distance}
Many default implementations of RF algorithms allow trees to grow to unrestricted depth \cite{breiman2001random}. As a consequence, on any given dataset, the trees in RF models tend to be deeper than those in GBT models. Table \ref{tab:tree_depth}, which presents statistics of tree depth in RF and GBT models, confirms this. The shallower the tree, the fewer leaves, and the higher the probability of a pair of points ending up in the same leaf. 
With larger datasets, conversely, RF trees tend to get deeper (Table \ref{tab:tree_depth}). However, the GBT trees we generate remain limited to depth 3 to 5. Hence, the larger the dataset, the more different we expect RF and GBT distances to be. 

\begin{table}[hb!]
\centering
\small
\begin{tabular}{lllllllll}
\toprule
 & & GBT & \multicolumn{4}{c}{RF Depth} \\
 \cmidrule{4-7}
Dataset & $n$ & Depth & Min & Mean & Max & Var  \\
\midrule
Breastcancer & 569 & 3 & 2 & 3.02 & 4 & 0.40  \\
Diabetes & 768 & 3 & 5 & 7.36 & 12 & 1.04 \\
T-COMPAS & 1000 & 4 & 6 & 8.70 & 14 & 1.18 \\
RHC & 5735 & 3 & 11& 14.7 & 21 & 1.41 \\
MNIST 4-9 & 5000 & 3 & 8 & 10.95 & 16 & 1.44 \\ 
CAL256 G-M & 215 & 2 & 2 & 2.51 & 3 & 0.50 \\ 
\bottomrule
\end{tabular}
\caption{Statistics of RF and GBT models tree depth across different datasets. $n$ is the number of observations in the dataset. All RF models had 1000 trees. All GBT models had an optimal number of trees (based on validation set loss) less than or equal to 200.}
\label{tab:tree_depth}
\end{table}

\begin{figure*}
     \centering
     \includegraphics[width=0.4\textwidth]{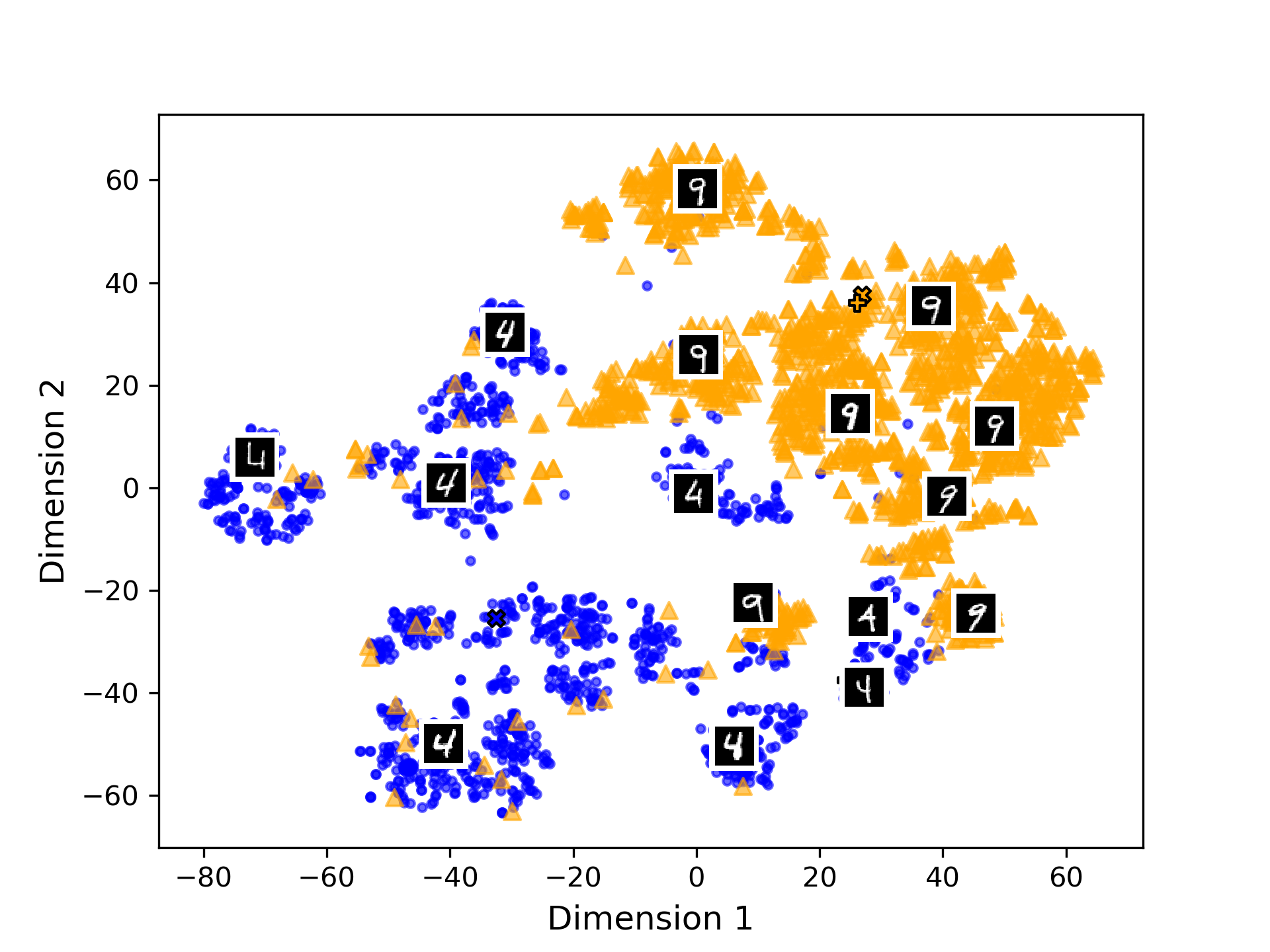}
     \includegraphics[width=0.4\textwidth]{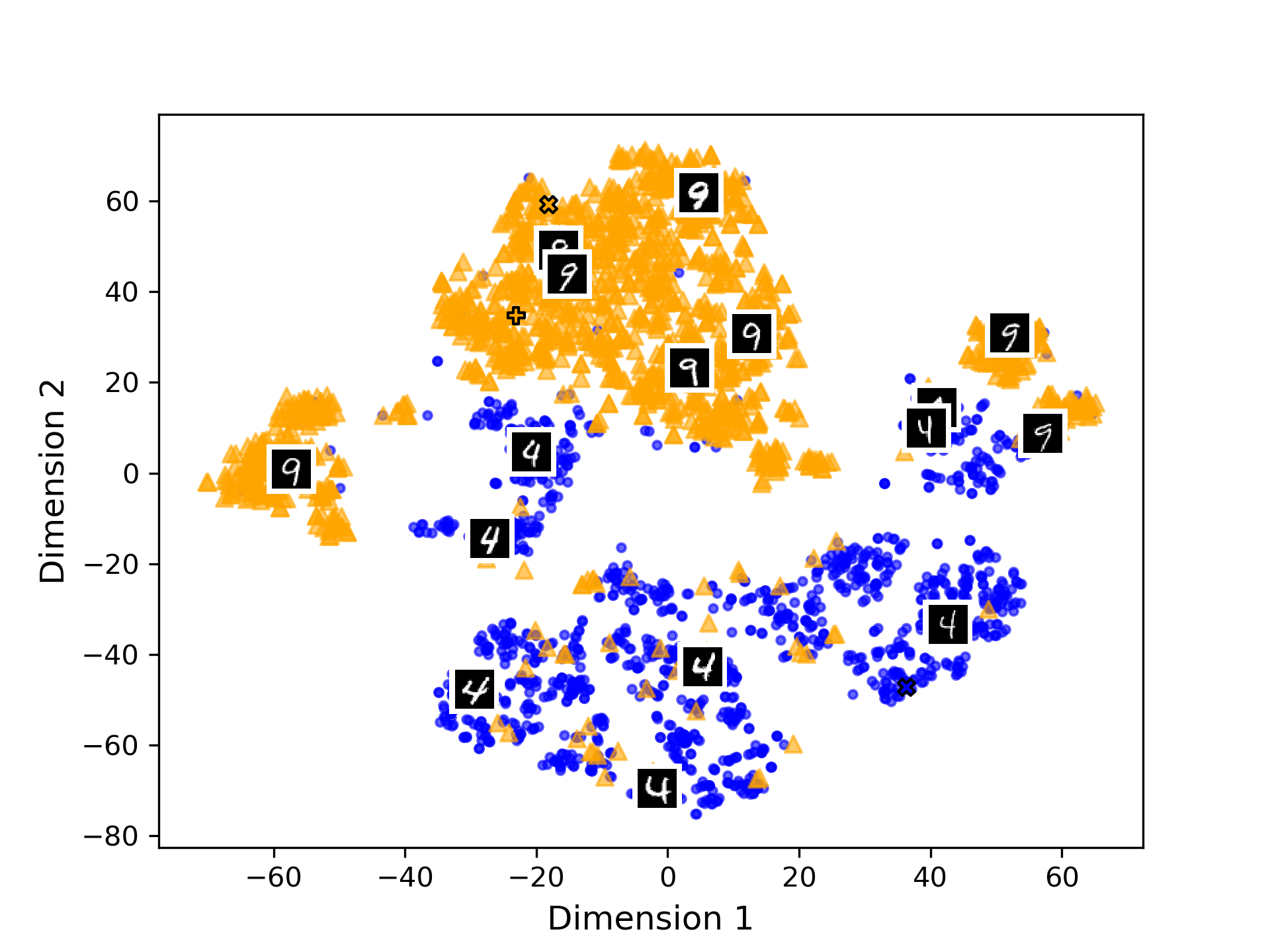}
    
     \includegraphics[width=0.4\textwidth]{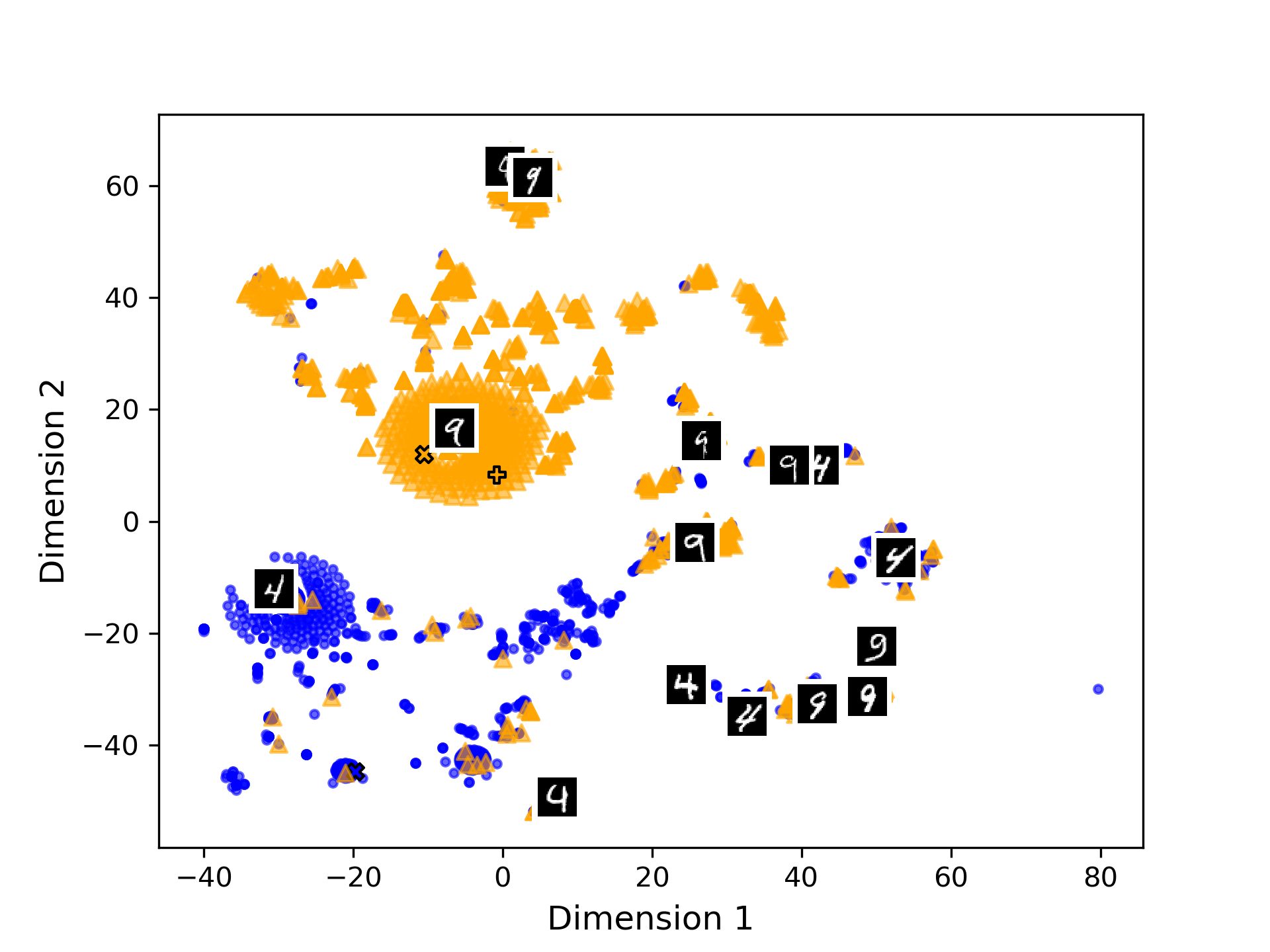}
     \includegraphics[width=0.4\textwidth]{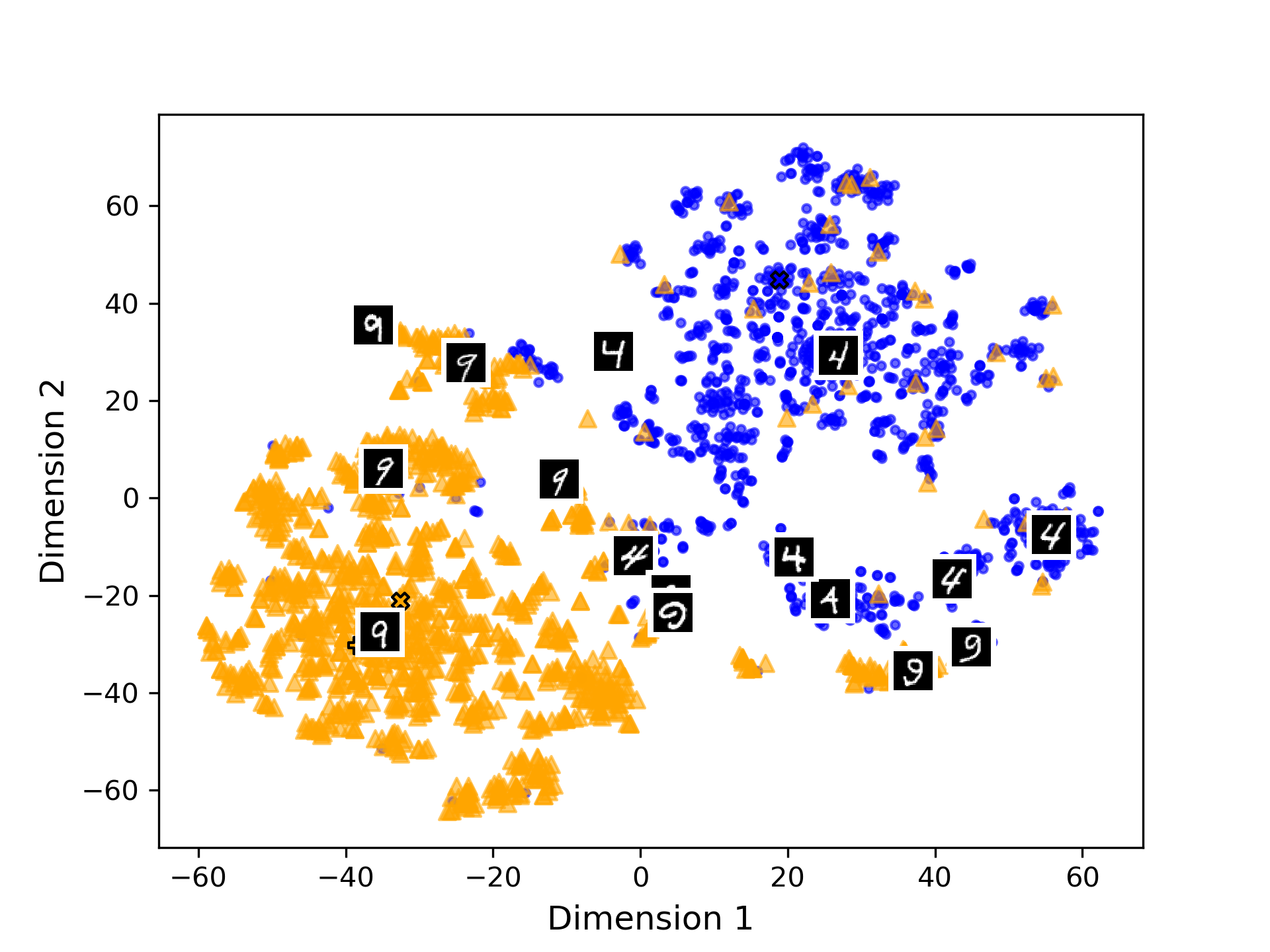}
    
     \caption{Visualization of distances using t-sne
 for optimal GBT with unweighted trees (top left), trees weighted by $vg^2$ (top right), RF with short trees matching GBT depth of 3 (bottom left), and optimal RF (bottom right) with mean depth 16 on the MNIST 4-9 dataset, using the \textcolor{greenmatplotlib}{adaptive greedy submodular (SM-A)} prototype selection method. \textcolor{orange}{Orange} represents the digit 9, \textcolor{blue}{blue} represents the digit 4. The two points marked by \texttt{x} and \texttt{+} are the same across all subfigures, to indicate how points move across different distance representations. Note that the bottom right subfigure is the same as the left subfigure in Figure \ref{fig:comparingoptimizersmnist}.}
     \label{fig:comparingmetrics}
 \end{figure*}

Figure \ref{fig:comparingmetrics} visualizes RF and GBT distances embedded in a two-dimensional space using t-sne and several MNIST prototypes. While digits 4 and 9 are clearly separable in Figure \ref{fig:comparingmetrics}, consistent with the high performance (97\% accuracy in Table \ref{tab:validated_prototype}) of the RF and GBT models, the models appear to be learning different representations, with GBT grouping points together in smaller and more compact clusters than RF. Accordingly, the prototypes selected for the RF distance are also different from those selected for GBT.

A natural next question may be the following: to what degree are differences between GBT and RF distances caused by different tree depth, different weights used in constructing the distance, or that different patterns in the data are being learned by RF compared to GBT models? While the top right corner of Figure \ref{fig:comparingmetrics} visualizes distances from GBT models trained with default settings (short), and the bottom right corner of the same figures depicts distances from RF models trained with default settings (unrestricted depth), the bottom left corner shows RF models trained \emph{to the same depth} as the corresponding default GBT model on that dataset. While the short RF model has smaller and more compact clusters than the default RF model, the RF and GBT models of same depth can still be told apart. 

Finally, the top left corner of Figure \ref{fig:comparingmetrics} visualizes an unweighted distance function derived from the same GBT model as in the top right corner, which uses a weighted distance function. This unweighted GBT distance (top left corner) looks more similar to the default RF model's distance (bottom right corner).

\subsection{Evaluating Interpretability: User Study}
To evaluate if prototypes are interpretable to humans -- the intended end users of our method -- we follow the human-grounded evaluation framework outlined by Doshi-Velez and Kim \cite{finale2017towards}, and design a user study where the task is to predict what the model would predict, after being presented with an explanation and inputs to the model. This task is exactly the ``forward simulation/prediction'' experiment described by Doshi-Velez and Kim \cite{finale2017towards}, and fulfills the simulatability criterion for interpretability, one of several criteria proposed by Lipton \cite{lipton2016mythos}. 

We compare prototypes to TreeExplainer \cite{lundberg2020local}, a Shapley values feature attribution method \cite{Lundberg2017unified} for tree ensembles. Here, we compare to Shapley values because it is a state-of-the-art feature attribution method that satisfies several axiomatic guarantees \cite{Lundberg2017unified}, is popular in practice \cite{bhatt2020explainable}, and, as a feature-based explanation, is presented differently to users. Hence, this comparison can inform us whether prototypes are indeed a viable alternative to feature-based explanations for tree ensembles. The hypothesis we investigate in this user study is therefore whether users are able to correctly predict a tree ensemble model's output using prototypes, and moreover, if they are able to do so with greater accuracy than when using Shapley values.

\subsubsection{User Study Design} We recruited participants on Amazon Mechanical Turk to participate in the study. We selected a dataset on vehicle fuel efficiency, from the R ggplot2 package (\url{https://ggplot2.tidyverse.org/reference/mpg.html}). The label is whether the vehicle is fuel efficient (greater than 19 highway miles per gallon), which does not require expert domain knowledge to understand. 

Each user was randomly assigned to either the prototypes or Shapley condition, with no users in both conditions. Users were presented with model inputs (Figure \ref{fig:userstudy_inputs} in the appendix); users in the prototype condition were presented with prototype explanations (Figure \ref{fig:userstudy_prot}) obtained by applying our prototype selection method to a tree ensemble; users in the Shapley condition were presented with a set of Shapley plots (Figure \ref{fig:userstudy_shap}) with added guidelines on how to interpret the plots. Then, users were presented a question (Figure \ref{fig:userstudy_questions}) and asked to predict what the model would predict for that vehicle. Each user was asked to evaluate 13 vehicles. These vehicles were randomly selected from the test set, while ensuring that every combination of 13 vehicles seen by a user in the prototype condition was also seen by another user in the Shapley condition, to account for certain vehicles being more difficult to predict according to either the model or human intuition.

To ensure that participants were paying attention and trying to answer the questions to the best of their ability, we designed two of these 13 questions to be ability and attention checks, known to help in identifying inattentive participants \cite{goodman2013data}. In particular, for the attention check, users were asked to select "fuel efficient" for a specific vehicle, regardless of what they believe the true answer is. Users were compensated \$3.00 upon completion of the study. After removing users who failed either catch trial, 42 users remained.

\subsubsection{User Study Results} We evaluate the results using a \textbf{human accuracy} metric: the fraction of vehicles where users predicted correctly the model's prediction for that vehicle. We also report user responses to a question at the end of the survey about how confident they felt about their answers. 

Table \ref{tab:userstudy} presents the results of two-sample t-tests comparing these metrics for the prototype and Shapley conditions. The p-values are one-sided, testing the alternative hypothesis if the metric is greater in the prototype condition than Shapley condition (i.e. higher accuracy, or greater confidence). The average self-reported confidence (on a scale of 1-4, higher is more confident) among users who used prototypes was 2.783, with users who used Shapley values self-reporting confidence of 2.736. This difference was not statistically significant. We note that self-reported user confidence has not been found to be indicative of actual performance and can sometimes even be misleading \cite{lakkaraju2020fool,green2019disparate}, with studies finding that humans cannot accurately assess their own performance \cite{lai2019human, poursabzi2018manipulating}. In contrast, there is a statistically significant improvement in human accuracy when using prototypes compared to Shapley values (0.79 compared to 0.72 human accuracy; p-value 0.035), demonstrating that prototypes are a viable alternative to feature-based explanation methods.

\begin{table}[hb!]
\centering
\small
\begin{tabular}{llll}
\toprule
\textbf{Metrics} & Prototypes & Shapley values & t-test p-value\\
\midrule
Human accuracy & 0.79 & 0.72 & 0.035* \\
Confidence & 2.783 & 2.736 & 0.238 \\
\bottomrule
\end{tabular}
\caption{Results from user study. Statistically significant differences are marked by *. The difference in human accuracy between prototype and Shapley conditions is statistically significant, with a one-sided t-test with alternative hypothesis that accuracy is higher in the prototype condition than Shapley condition returning p-value of 0.035.}
\label{tab:userstudy}
\end{table}

We also collected qualitative feedback on the explanations and user interface. When asked ``Was anything confusing? Is there anything you would have liked to know that would have helped you better answer these questions?'' at the end of the survey, one user in the Shapley values condition responded ``It would have helped if [Shapley values] showed the difference better'', and another user reported ``The way the values were weighted seemed a bit strange to me''. On the other hand, in response to the same question, one user in the prototypes condition was able to articulate a simple mental model of what s/he thought the model was doing, saying ``Knowing the make and model of the vehicle would have been helpful, but maybe I think that because I'm familiar with cars. (...) [The model] seems to think that vehicles with smaller engines will automatically be more fuel efficient, which is why I made the choices that I did because the instructions said to guess the predictions of the model.``.

\subsection{A Use Case of Fixing Mislabeled Points}
We now demonstrate how the ideas presented in this paper can assist with certain tasks that machine learning model users may wish to leverage explanations for, such as debugging a dataset. We focus on the specific task of correcting mislabeled points in a dataset, where we wish to present a reasonable number of points to the user, in some meaningful ordering, for the user to correct any wrong labels. This experiment is similar to that ran by \cite{koh2017understanding,yeh2018representer}.

We corrupt the MNIST 4-9 dataset by flipping the labels of 33\% of points, and use RF distance to construct a ranking of points, which we compare to two baselines: (1) random ranking; (2) ranking based on loss of training points. The distance ranking is constructed thus: for every training point, compute $k=10$ nearest neighbors based on the distance, then compute the proportion of neighbors that share the same label as the point. 
Note that the loss ranking is strong baseline, as found by \cite{koh2017understanding}.

\begin{figure}[ht!]
\centering
    \includegraphics[width=0.35\textwidth]{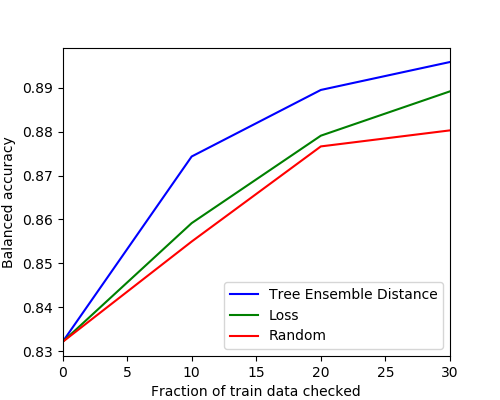}
    \caption{Performance of different algorithms, including one based on tree ensemble distance, to select points from MNIST 4-9 with corrupted labels for a simulated user to inspect. The simulated user flips the labels of points checked, and a model is retrained on the corrected data.}
    \label{fig:corrupt}
\end{figure}

We present a simulated user with a proportion (up to 30\%) of training points, as ranked by the different methods, to inspect and correct. Similar to \cite{koh2017understanding,yeh2018representer}, the simulated user is an oracle who only corrects points that are flipped, of all the points presented to her. The model is then retrained on the corrected data. We repeat the experiment 20 times, randomizing the points corrupted each time, and average the results. Figure \ref{fig:corrupt} plots the mean test set performance of the model retrained on simulated user corrected data. With the same interpretability budget (amount of points the simulated user had to inspect), tree ensemble distance based ranking was better at assisting in correcting mislabeled points, generating corrected data that had higher test set accuracy than other rankings.

\section{Concluding remarks}

We have proposed that tree ensemble classifiers can be made interpretable using prototypes that are central according to a distance function derived from the tree ensemble. Our user study suggests that this is indeed the case, as humans were better at predicting the output of a tree ensemble classifier using prototypes than Shapley values, a popular feature-based approach. At the same time, our quantitative evaluation (Table \ref{tab:validated_prototype}) suggests that in many cases, tree space prototypes are able to capture much of the power of the original classifier, or sometimes even constitute a superior separate classifier, since these prototypes when used as a nearest-prototype classifier sometimes outperform the original tree ensemble. 
This suggests that to the extent that tree ensemble proximity is an intuitive measure of similarity, 
users with longer-term experience with prototypes may attain even better accuracy at anticipating the behavior of the tree ensemble.
Another intriguing possibility is that we ought to consider a slightly different question: 
if a nearest-prototype classifier is the superior classifier, should we perhaps analyse its interpretability in its own right, 
as opposed to the original tree ensemble?

Another open question is the impact of the number of prototypes on interpretability, as opposed to accuracy alone. Intuition suggests that humans should be better at dealing with a smaller number of prototypes, but how does this trade off against the generally greater accuracy nearest-prototype classifiers tend to exhibit with more prototypes? Subjectively, our adaptive prototype selection methods that can select different numbers of prototypes per class tend to lead to better coverage for classes that are more diverse, but our analysis shows that there do exist some datasets on which they do not benefit classification. A quantitative analysis of the relationship between prototype count and human predictions in multiple settings could provide further arguments for or against adaptive methods, and constitutes an attractive direction for future work.

\begin{acks}
We thank Jacob Bien and Albert Gordo for helpful discussion. GH was supported by NSF grant DMS-1712554. MTW was supported by NIH grants R01 GM135926 and U19 AI111143.
\end{acks}

\bibliographystyle{ACM-Reference-Format}
\bibliography{main}


\begin{thebibliography}{54}


\ifx \showCODEN    \undefined \def \showCODEN     #1{\unskip}     \fi
\ifx \showDOI      \undefined \def \showDOI       #1{#1}\fi
\ifx \showISBNx    \undefined \def \showISBNx     #1{\unskip}     \fi
\ifx \showISBNxiii \undefined \def \showISBNxiii  #1{\unskip}     \fi
\ifx \showISSN     \undefined \def \showISSN      #1{\unskip}     \fi
\ifx \showLCCN     \undefined \def \showLCCN      #1{\unskip}     \fi
\ifx \shownote     \undefined \def \shownote      #1{#1}          \fi
\ifx \showarticletitle \undefined \def \showarticletitle #1{#1}   \fi
\ifx \showURL      \undefined \def \showURL       {\relax}        \fi
\providecommand\bibfield[2]{#2}
\providecommand\bibinfo[2]{#2}
\providecommand\natexlab[1]{#1}
\providecommand\showeprint[2][]{arXiv:#2}

\bibitem[\protect\citeauthoryear{Bhatt, Xiang, Sharma, Weller, Taly, Jia,
  Ghosh, Puri, Moura, and Eckersley}{Bhatt et~al\mbox{.}}{2020}]%
        {bhatt2020explainable}
\bibfield{author}{\bibinfo{person}{Umang Bhatt}, \bibinfo{person}{Alice Xiang},
  \bibinfo{person}{Shubham Sharma}, \bibinfo{person}{Adrian Weller},
  \bibinfo{person}{Ankur Taly}, \bibinfo{person}{Yunhan Jia},
  \bibinfo{person}{Joydeep Ghosh}, \bibinfo{person}{Ruchir Puri},
  \bibinfo{person}{Jose~MF Moura}, {and} \bibinfo{person}{Peter Eckersley}.}
  \bibinfo{year}{2020}\natexlab{}.
\newblock \showarticletitle{Explainable machine learning in deployment}. In
  \bibinfo{booktitle}{\emph{FAT*}}.
\newblock


\bibitem[\protect\citeauthoryear{Bien and Tibshirani}{Bien and
  Tibshirani}{2011}]%
        {bien2011prototype}
\bibfield{author}{\bibinfo{person}{Jacob Bien} {and} \bibinfo{person}{Robert
  Tibshirani}.} \bibinfo{year}{2011}\natexlab{}.
\newblock \showarticletitle{Prototype selection for interpretable
  classification}.
\newblock \bibinfo{journal}{\emph{The Annals of Applied Statistics}}
  (\bibinfo{year}{2011}).
\newblock


\bibitem[\protect\citeauthoryear{Breiman}{Breiman}{2001}]%
        {breiman2001random}
\bibfield{author}{\bibinfo{person}{Leo Breiman}.}
  \bibinfo{year}{2001}\natexlab{}.
\newblock \showarticletitle{Random forests}.
\newblock \bibinfo{journal}{\emph{Machine Learning}} \bibinfo{volume}{45},
  \bibinfo{number}{1} (\bibinfo{year}{2001}).
\newblock


\bibitem[\protect\citeauthoryear{Breiman and Cutler}{Breiman and
  Cutler}{2002}]%
        {breiman2002rf}
\bibfield{author}{\bibinfo{person}{Leo Breiman} {and} \bibinfo{person}{Adele
  Cutler}.} \bibinfo{year}{2002}\natexlab{}.
\newblock \bibinfo{title}{Random Forests Manual}.
\newblock
  \bibinfo{howpublished}{\url{https://www.stat.berkeley.edu/~breiman/RandomForests}}.
\newblock
\newblock
\shownote{Accessed July 6, 2019. Year 2002 based on copyright year indicated in
  the authors' Fortran code.}


\bibitem[\protect\citeauthoryear{Brodersen, Ong, Stephan, and
  Buhmann}{Brodersen et~al\mbox{.}}{2010}]%
        {balancedaccuracy}
\bibfield{author}{\bibinfo{person}{Kay~Henning Brodersen},
  \bibinfo{person}{Cheng~Soon Ong}, \bibinfo{person}{Klaas~Enno Stephan}, {and}
  \bibinfo{person}{Joachim~M Buhmann}.} \bibinfo{year}{2010}\natexlab{}.
\newblock \showarticletitle{The Balanced Accuracy and Its Posterior
  Distribution}. In \bibinfo{booktitle}{\emph{International Conference on
  Pattern Recognition}}.
\newblock


\bibitem[\protect\citeauthoryear{Caruana, Kangarloo, Dionisio, Sinha, and
  Johnson}{Caruana et~al\mbox{.}}{1999}]%
        {caruana1999case}
\bibfield{author}{\bibinfo{person}{Rich Caruana}, \bibinfo{person}{Hooshang
  Kangarloo}, \bibinfo{person}{John~David Dionisio}, \bibinfo{person}{Usha
  Sinha}, {and} \bibinfo{person}{David Johnson}.}
  \bibinfo{year}{1999}\natexlab{}.
\newblock \showarticletitle{Case-based explanation of non-case-based learning
  methods.}. In \bibinfo{booktitle}{\emph{Proceedings of the AMIA Symposium}}.
  American Medical Informatics Association.
\newblock


\bibitem[\protect\citeauthoryear{Caruana and Niculescu-Mizil}{Caruana and
  Niculescu-Mizil}{2006}]%
        {caruana2006empirical}
\bibfield{author}{\bibinfo{person}{Rich Caruana} {and}
  \bibinfo{person}{Alexandru Niculescu-Mizil}.}
  \bibinfo{year}{2006}\natexlab{}.
\newblock \showarticletitle{An Empirical Comparison of Supervised Learning
  Algorithms}. In \bibinfo{booktitle}{\emph{ICML}}.
\newblock


\bibitem[\protect\citeauthoryear{Doshi-Velez and Kim}{Doshi-Velez and
  Kim}{2017}]%
        {finale2017towards}
\bibfield{author}{\bibinfo{person}{Finale Doshi-Velez} {and}
  \bibinfo{person}{Been Kim}.} \bibinfo{year}{2017}\natexlab{}.
\newblock \showarticletitle{Towards A Rigorous Science of Interpretable Machine
  Learning}.
\newblock \bibinfo{journal}{\emph{arXiv preprint arXiv:1702.08608}}
  (\bibinfo{year}{2017}).
\newblock


\bibitem[\protect\citeauthoryear{Dressel and Farid}{Dressel and Farid}{2018}]%
        {dressel2018accuracy}
\bibfield{author}{\bibinfo{person}{Julia Dressel} {and} \bibinfo{person}{Hany
  Farid}.} \bibinfo{year}{2018}\natexlab{}.
\newblock \showarticletitle{The accuracy, fairness, and limits of predicting
  recidivism}.
\newblock \bibinfo{journal}{\emph{Science Advances}} (\bibinfo{year}{2018}).
\newblock
\newblock
\shownote{Data accessed from
  \url{www.cs.dartmouth.edu/farid/downloads/publications/scienceadvances17}.}


\bibitem[\protect\citeauthoryear{Freitas}{Freitas}{2014}]%
        {freitas2014comprehensible}
\bibfield{author}{\bibinfo{person}{Alex~A Freitas}.}
  \bibinfo{year}{2014}\natexlab{}.
\newblock \showarticletitle{Comprehensible classification models: a position
  paper}.
\newblock \bibinfo{journal}{\emph{ACM SIGKDD Explorations}}
  (\bibinfo{year}{2014}).
\newblock


\bibitem[\protect\citeauthoryear{Friedman}{Friedman}{2001}]%
        {friedman2001greedy}
\bibfield{author}{\bibinfo{person}{Jerome~H Friedman}.}
  \bibinfo{year}{2001}\natexlab{}.
\newblock \showarticletitle{Greedy function approximation: a gradient boosting
  machine}.
\newblock \bibinfo{journal}{\emph{Annals of Statistics}}
  (\bibinfo{year}{2001}).
\newblock


\bibitem[\protect\citeauthoryear{Garcia, Derrac, Cano, and Herrera}{Garcia
  et~al\mbox{.}}{2011}]%
        {garcia2011prototype}
\bibfield{author}{\bibinfo{person}{Salvador Garcia}, \bibinfo{person}{Joaquin
  Derrac}, \bibinfo{person}{Jose~Ramon Cano}, {and} \bibinfo{person}{Francisco
  Herrera}.} \bibinfo{year}{2011}\natexlab{}.
\newblock \showarticletitle{Prototype selection for nearest neighbor
  classification: Taxonomy and empirical study}.
\newblock \bibinfo{journal}{\emph{TPAMI}} (\bibinfo{year}{2011}).
\newblock


\bibitem[\protect\citeauthoryear{Gomes and Krause}{Gomes and Krause}{2010}]%
        {gomeskrause}
\bibfield{author}{\bibinfo{person}{Ryan Gomes} {and} \bibinfo{person}{Andreas
  Krause}.} \bibinfo{year}{2010}\natexlab{}.
\newblock \showarticletitle{Budgeted Nonparametric Learning from Data Streams}.
  In \bibinfo{booktitle}{\emph{ICML}}.
\newblock


\bibitem[\protect\citeauthoryear{Goodman, Cryder, and Cheema}{Goodman
  et~al\mbox{.}}{2013}]%
        {goodman2013data}
\bibfield{author}{\bibinfo{person}{Joseph~K Goodman},
  \bibinfo{person}{Cynthia~E Cryder}, {and} \bibinfo{person}{Amar Cheema}.}
  \bibinfo{year}{2013}\natexlab{}.
\newblock \showarticletitle{Data Collection in a Flat World: The Strengths and
  Weaknesses of Mechanical Turk Samples}.
\newblock \bibinfo{journal}{\emph{Journal of Behavioral Decision Making}}
  \bibinfo{volume}{26} (\bibinfo{year}{2013}).
\newblock


\bibitem[\protect\citeauthoryear{Green and Chen}{Green and Chen}{2019}]%
        {green2019disparate}
\bibfield{author}{\bibinfo{person}{Ben Green} {and} \bibinfo{person}{Yiling
  Chen}.} \bibinfo{year}{2019}\natexlab{}.
\newblock \showarticletitle{Disparate interactions: An algorithm-in-the-loop
  analysis of fairness in risk assessments}. In
  \bibinfo{booktitle}{\emph{FAT*}}.
\newblock


\bibitem[\protect\citeauthoryear{Griffin, Holub, and Perona}{Griffin
  et~al\mbox{.}}{2006}]%
        {caltech256}
\bibfield{author}{\bibinfo{person}{Gregory Griffin}, \bibinfo{person}{Alex
  Holub}, {and} \bibinfo{person}{Pietro Perona}.}
  \bibinfo{year}{2006}\natexlab{}.
\newblock \showarticletitle{Caltech-256 Object Category Dataset}.
\newblock \bibinfo{journal}{\emph{Technical Report, California Institute of
  Technology}} (\bibinfo{year}{2006}).
\newblock


\bibitem[\protect\citeauthoryear{Hadsell, Chopra, and LeCun}{Hadsell
  et~al\mbox{.}}{2006}]%
        {hadsell2006dimensionality}
\bibfield{author}{\bibinfo{person}{Raia Hadsell}, \bibinfo{person}{Sumit
  Chopra}, {and} \bibinfo{person}{Yann LeCun}.}
  \bibinfo{year}{2006}\natexlab{}.
\newblock \showarticletitle{Dimensionality Reduction by Learning an Invariant
  Mapping}. In \bibinfo{booktitle}{\emph{CVPR}}.
\newblock


\bibitem[\protect\citeauthoryear{Hara and Hayashi}{Hara and Hayashi}{2018}]%
        {hara2016making}
\bibfield{author}{\bibinfo{person}{Satoshi Hara} {and} \bibinfo{person}{Kohei
  Hayashi}.} \bibinfo{year}{2018}\natexlab{}.
\newblock \showarticletitle{Making tree ensembles interpretable: A Bayesian
  Model Selection Approach}. In \bibinfo{booktitle}{\emph{AISTATS}}.
\newblock


\bibitem[\protect\citeauthoryear{Hart}{Hart}{1968}]%
        {hart1968condensed}
\bibfield{author}{\bibinfo{person}{Peter Hart}.}
  \bibinfo{year}{1968}\natexlab{}.
\newblock \showarticletitle{The condensed nearest neighbor rule}.
\newblock \bibinfo{journal}{\emph{IEEE Transactions on Information Theory}}
  (\bibinfo{year}{1968}).
\newblock


\bibitem[\protect\citeauthoryear{Ishwaran}{Ishwaran}{2007}]%
        {ishwaran2007variable}
\bibfield{author}{\bibinfo{person}{Hemant Ishwaran}.}
  \bibinfo{year}{2007}\natexlab{}.
\newblock \showarticletitle{Variable importance in binary regression trees and
  forests}.
\newblock \bibinfo{journal}{\emph{Electronic Journal of Statistics}}
  (\bibinfo{year}{2007}).
\newblock


\bibitem[\protect\citeauthoryear{Kaufman and Rousseeuw}{Kaufman and
  Rousseeuw}{1987}]%
        {kaufman1987clustering}
\bibfield{author}{\bibinfo{person}{Leonard Kaufman} {and}
  \bibinfo{person}{Peter~J Rousseeuw}.} \bibinfo{year}{1987}\natexlab{}.
\newblock \showarticletitle{Clustering by means of medoids}.
\newblock In \bibinfo{booktitle}{\emph{Statistical Data Analysis Based on the
  L1 Norm}}. \bibinfo{publisher}{Birkhäuser Basel}.
\newblock


\bibitem[\protect\citeauthoryear{Khanna, Kim, Ghosh, and Koyejo}{Khanna
  et~al\mbox{.}}{2019}]%
        {khanna2018interpreting}
\bibfield{author}{\bibinfo{person}{Rajiv Khanna}, \bibinfo{person}{Been Kim},
  \bibinfo{person}{Joydeep Ghosh}, {and} \bibinfo{person}{Oluwasanmi Koyejo}.}
  \bibinfo{year}{2019}\natexlab{}.
\newblock \showarticletitle{Interpreting black box predictions using fisher
  kernels}. In \bibinfo{booktitle}{\emph{AISTATS}}.
\newblock


\bibitem[\protect\citeauthoryear{Kim, Khanna, and Koyejo}{Kim
  et~al\mbox{.}}{2016}]%
        {kim2016examples}
\bibfield{author}{\bibinfo{person}{Been Kim}, \bibinfo{person}{Rajiv Khanna},
  {and} \bibinfo{person}{Oluwasanmi~O Koyejo}.}
  \bibinfo{year}{2016}\natexlab{}.
\newblock \showarticletitle{Examples are not enough, learn to criticize!
  criticism for interpretability}. In \bibinfo{booktitle}{\emph{NIPS}}.
\newblock


\bibitem[\protect\citeauthoryear{Kim, Rudin, and Shah}{Kim
  et~al\mbox{.}}{2014}]%
        {kim2014bayesian}
\bibfield{author}{\bibinfo{person}{Been Kim}, \bibinfo{person}{Cynthia Rudin},
  {and} \bibinfo{person}{Julie~A Shah}.} \bibinfo{year}{2014}\natexlab{}.
\newblock \showarticletitle{The Bayesian Case Model: A generative approach for
  case-based reasoning and prototype classification}. In
  \bibinfo{booktitle}{\emph{NIPS}}.
\newblock


\bibitem[\protect\citeauthoryear{Koh and Liang}{Koh and Liang}{2017}]%
        {koh2017understanding}
\bibfield{author}{\bibinfo{person}{Pang~Wei Koh} {and} \bibinfo{person}{Percy
  Liang}.} \bibinfo{year}{2017}\natexlab{}.
\newblock \showarticletitle{Understanding black-box predictions via influence
  functions}. In \bibinfo{booktitle}{\emph{ICML}}.
\newblock


\bibitem[\protect\citeauthoryear{Lai and Tan}{Lai and Tan}{2019}]%
        {lai2019human}
\bibfield{author}{\bibinfo{person}{Vivian Lai} {and} \bibinfo{person}{Chenhao
  Tan}.} \bibinfo{year}{2019}\natexlab{}.
\newblock \showarticletitle{On human predictions with explanations and
  predictions of machine learning models: A case study on deception detection}.
  In \bibinfo{booktitle}{\emph{FAT*}}.
\newblock


\bibitem[\protect\citeauthoryear{Lakkaraju and Bastani}{Lakkaraju and
  Bastani}{2020}]%
        {lakkaraju2020fool}
\bibfield{author}{\bibinfo{person}{Himabindu Lakkaraju} {and}
  \bibinfo{person}{Osbert Bastani}.} \bibinfo{year}{2020}\natexlab{}.
\newblock \showarticletitle{" How do I fool you?" Manipulating User Trust via
  Misleading Black Box Explanations}. In \bibinfo{booktitle}{\emph{AIES}}.
\newblock


\bibitem[\protect\citeauthoryear{Larson, Mattu, Kirchner, and Angwin}{Larson
  et~al\mbox{.}}{2016}]%
        {propublica2018compas}
\bibfield{author}{\bibinfo{person}{Jeff Larson}, \bibinfo{person}{Surya Mattu},
  \bibinfo{person}{Lauren Kirchner}, {and} \bibinfo{person}{Julia Angwin}.}
  \bibinfo{year}{2016}\natexlab{}.
\newblock \bibinfo{title}{How We Analyzed the COMPAS Recidivism Algorithm}.
\newblock \bibinfo{howpublished}{ProPublica}.
\newblock


\bibitem[\protect\citeauthoryear{LeCun, Bottou, Bengio, and Haffner}{LeCun
  et~al\mbox{.}}{1998}]%
        {mnist}
\bibfield{author}{\bibinfo{person}{Yann LeCun}, \bibinfo{person}{Leon Bottou},
  \bibinfo{person}{Yoshua Bengio}, {and} \bibinfo{person}{Patrick Haffner}.}
  \bibinfo{year}{1998}\natexlab{}.
\newblock \showarticletitle{Gradient-based learning applied to document
  recognition}. In \bibinfo{booktitle}{\emph{Proceedings of the IEEE}}.
\newblock


\bibitem[\protect\citeauthoryear{Li, Liu, Chen, and Rudin}{Li
  et~al\mbox{.}}{2018}]%
        {li2018deep}
\bibfield{author}{\bibinfo{person}{Oscar Li}, \bibinfo{person}{Hao Liu},
  \bibinfo{person}{Chaofan Chen}, {and} \bibinfo{person}{Cynthia Rudin}.}
  \bibinfo{year}{2018}\natexlab{}.
\newblock \showarticletitle{Deep learning for case-based reasoning through
  prototypes: A neural network that explains its predictions}. In
  \bibinfo{booktitle}{\emph{AAAI}}.
\newblock


\bibitem[\protect\citeauthoryear{Liaw and Wiener}{Liaw and Wiener}{2002}]%
        {liaw2002classification}
\bibfield{author}{\bibinfo{person}{Andy Liaw} {and} \bibinfo{person}{Matthew
  Wiener}.} \bibinfo{year}{2002}\natexlab{}.
\newblock \showarticletitle{Classification and regression by randomForest}.
\newblock \bibinfo{journal}{\emph{R News}} (\bibinfo{year}{2002}).
\newblock


\bibitem[\protect\citeauthoryear{Lin and Bilmes}{Lin and Bilmes}{2011}]%
        {lin2011class}
\bibfield{author}{\bibinfo{person}{Hui Lin} {and} \bibinfo{person}{Jeff
  Bilmes}.} \bibinfo{year}{2011}\natexlab{}.
\newblock \showarticletitle{A class of submodular functions for document
  summarization}. In \bibinfo{booktitle}{\emph{ACL}}.
\newblock


\bibitem[\protect\citeauthoryear{Lin and Jeon}{Lin and Jeon}{2006}]%
        {lin2006random}
\bibfield{author}{\bibinfo{person}{Yi Lin} {and} \bibinfo{person}{Yongho
  Jeon}.} \bibinfo{year}{2006}\natexlab{}.
\newblock \showarticletitle{Random forests and adaptive nearest neighbors}.
\newblock \bibinfo{journal}{\emph{J. Amer. Statist. Assoc.}}
  (\bibinfo{year}{2006}).
\newblock


\bibitem[\protect\citeauthoryear{Lipton}{Lipton}{2016}]%
        {lipton2016mythos}
\bibfield{author}{\bibinfo{person}{Zachary~C. Lipton}.}
  \bibinfo{year}{2016}\natexlab{}.
\newblock \showarticletitle{The Mythos of Model Interpretability}.
\newblock \bibinfo{journal}{\emph{arXiv preprint arXiv:1606.03490}}
  (\bibinfo{year}{2016}).
\newblock


\bibitem[\protect\citeauthoryear{Louppe}{Louppe}{2014}]%
        {louppe2014understanding}
\bibfield{author}{\bibinfo{person}{Gilles Louppe}.}
  \bibinfo{year}{2014}\natexlab{}.
\newblock \showarticletitle{Understanding random forests: From theory to
  practice}.
\newblock \bibinfo{journal}{\emph{arXiv preprint arXiv:1407.7502}}
  (\bibinfo{year}{2014}).
\newblock


\bibitem[\protect\citeauthoryear{Lundberg, Erion, Chen, DeGrave, Prutkin, Nair,
  Katz, Himmelfarb, Bansal, and Lee}{Lundberg et~al\mbox{.}}{2020}]%
        {lundberg2020local}
\bibfield{author}{\bibinfo{person}{Scott~M Lundberg}, \bibinfo{person}{Gabriel
  Erion}, \bibinfo{person}{Hugh Chen}, \bibinfo{person}{Alex DeGrave},
  \bibinfo{person}{Jordan~M Prutkin}, \bibinfo{person}{Bala Nair},
  \bibinfo{person}{Ronit Katz}, \bibinfo{person}{Jonathan Himmelfarb},
  \bibinfo{person}{Nisha Bansal}, {and} \bibinfo{person}{Su-In Lee}.}
  \bibinfo{year}{2020}\natexlab{}.
\newblock \showarticletitle{From local explanations to global understanding
  with explainable AI for trees}.
\newblock \bibinfo{journal}{\emph{Nature Machine Intelligence}}
  \bibinfo{volume}{2} (\bibinfo{year}{2020}).
\newblock


\bibitem[\protect\citeauthoryear{Lundberg and Lee}{Lundberg and Lee}{2017}]%
        {Lundberg2017unified}
\bibfield{author}{\bibinfo{person}{Scott~M Lundberg} {and}
  \bibinfo{person}{Su-In Lee}.} \bibinfo{year}{2017}\natexlab{}.
\newblock \showarticletitle{A Unified Approach to Interpreting Model
  Predictions}. In \bibinfo{booktitle}{\emph{NIPS}}.
\newblock


\bibitem[\protect\citeauthoryear{Maaten and Hinton}{Maaten and Hinton}{2008}]%
        {maaten2008visualizing}
\bibfield{author}{\bibinfo{person}{Laurens van~der Maaten} {and}
  \bibinfo{person}{Geoffrey Hinton}.} \bibinfo{year}{2008}\natexlab{}.
\newblock \showarticletitle{Visualizing data using t-SNE}.
\newblock \bibinfo{journal}{\emph{JMLR}} (\bibinfo{year}{2008}).
\newblock


\bibitem[\protect\citeauthoryear{Mirzasoleiman, Karbasi, Sarkar, and
  Krause}{Mirzasoleiman et~al\mbox{.}}{2013}]%
        {mirzasoleiman2013distributed}
\bibfield{author}{\bibinfo{person}{Baharan Mirzasoleiman},
  \bibinfo{person}{Amin Karbasi}, \bibinfo{person}{Rik Sarkar}, {and}
  \bibinfo{person}{Andreas Krause}.} \bibinfo{year}{2013}\natexlab{}.
\newblock \showarticletitle{Distributed submodular maximization: Identifying
  representative elements in massive data}. In
  \bibinfo{booktitle}{\emph{NIPS}}.
\newblock


\bibitem[\protect\citeauthoryear{Nemhauser, Wolsey, and Fisher}{Nemhauser
  et~al\mbox{.}}{1978}]%
        {nemhauser1978analysis}
\bibfield{author}{\bibinfo{person}{George~L Nemhauser},
  \bibinfo{person}{Laurence~A Wolsey}, {and} \bibinfo{person}{Marshall~L
  Fisher}.} \bibinfo{year}{1978}\natexlab{}.
\newblock \showarticletitle{An analysis of approximations for maximizing
  submodular set functions—I}.
\newblock \bibinfo{journal}{\emph{Mathematical Programming}}
  (\bibinfo{year}{1978}).
\newblock


\bibitem[\protect\citeauthoryear{Papadimitriou}{Papadimitriou}{1981}]%
        {papad1981}
\bibfield{author}{\bibinfo{person}{Christos~H Papadimitriou}.}
  \bibinfo{year}{1981}\natexlab{}.
\newblock \showarticletitle{Worst-Case and Probabilistic Analysis of a
  Geometric Location Problem}.
\newblock \bibinfo{journal}{\emph{SIAM J. Comput.}} (\bibinfo{year}{1981}).
\newblock


\bibitem[\protect\citeauthoryear{Poursabzi-Sangdeh, Goldstein, Hofman, Vaughan,
  and Wallach}{Poursabzi-Sangdeh et~al\mbox{.}}{2018}]%
        {poursabzi2018manipulating}
\bibfield{author}{\bibinfo{person}{Forough Poursabzi-Sangdeh},
  \bibinfo{person}{Daniel~G Goldstein}, \bibinfo{person}{Jake~M Hofman},
  \bibinfo{person}{Jennifer~Wortman Vaughan}, {and} \bibinfo{person}{Hanna
  Wallach}.} \bibinfo{year}{2018}\natexlab{}.
\newblock \showarticletitle{Manipulating and measuring model interpretability}.
\newblock \bibinfo{journal}{\emph{arXiv preprint arXiv:1802.07810}}
  (\bibinfo{year}{2018}).
\newblock


\bibitem[\protect\citeauthoryear{Ribeiro, Singh, and Guestrin}{Ribeiro
  et~al\mbox{.}}{2016}]%
        {lime}
\bibfield{author}{\bibinfo{person}{Marco~Tulio Ribeiro},
  \bibinfo{person}{Sameer Singh}, {and} \bibinfo{person}{Carlos Guestrin}.}
  \bibinfo{year}{2016}\natexlab{}.
\newblock \showarticletitle{{``Why Should I Trust You?"}: Explaining the
  Predictions of Any Classifier}. In \bibinfo{booktitle}{\emph{KDD}}.
\newblock


\bibitem[\protect\citeauthoryear{Richter and Aamodt}{Richter and
  Aamodt}{2005}]%
        {richter2005case}
\bibfield{author}{\bibinfo{person}{Michael~M Richter} {and}
  \bibinfo{person}{Agnar Aamodt}.} \bibinfo{year}{2005}\natexlab{}.
\newblock \showarticletitle{Case-based reasoning foundations}.
\newblock \bibinfo{journal}{\emph{The Knowledge Engineering Review}}
  (\bibinfo{year}{2005}).
\newblock


\bibitem[\protect\citeauthoryear{Scornet}{Scornet}{2016}]%
        {scornet2016random}
\bibfield{author}{\bibinfo{person}{Erwan Scornet}.}
  \bibinfo{year}{2016}\natexlab{}.
\newblock \showarticletitle{Random forests and kernel methods}.
\newblock \bibinfo{journal}{\emph{IEEE Transactions on Information Theory}}
  (\bibinfo{year}{2016}).
\newblock


\bibitem[\protect\citeauthoryear{Shi and Horvath}{Shi and Horvath}{2006}]%
        {shi2006unsupervised}
\bibfield{author}{\bibinfo{person}{Tao Shi} {and} \bibinfo{person}{Steve
  Horvath}.} \bibinfo{year}{2006}\natexlab{}.
\newblock \showarticletitle{Unsupervised learning with random forest
  predictors}.
\newblock \bibinfo{journal}{\emph{Journal of Computational and Graphical
  Statistics}} (\bibinfo{year}{2006}).
\newblock


\bibitem[\protect\citeauthoryear{Stekhoven}{Stekhoven}{2015}]%
        {stekhoven2015missforest}
\bibfield{author}{\bibinfo{person}{Daniel~J Stekhoven}.}
  \bibinfo{year}{2015}\natexlab{}.
\newblock \showarticletitle{missForest: Nonparametric missing value imputation
  using random forest}.
\newblock \bibinfo{journal}{\emph{Astrophysics Source Code Library}}
  (\bibinfo{year}{2015}).
\newblock


\bibitem[\protect\citeauthoryear{Wachter, Mittelstadt, and Russell}{Wachter
  et~al\mbox{.}}{2017}]%
        {wachter2017counterfactual}
\bibfield{author}{\bibinfo{person}{Sandra Wachter}, \bibinfo{person}{Brent
  Mittelstadt}, {and} \bibinfo{person}{Chris Russell}.}
  \bibinfo{year}{2017}\natexlab{}.
\newblock \showarticletitle{Counterfactual explanations without opening the
  black box: Automated decisions and the GDPR}.
\newblock \bibinfo{journal}{\emph{Harvard Journal of Law \& Technology}}
  \bibinfo{volume}{31}, \bibinfo{number}{2} (\bibinfo{year}{2017}).
\newblock


\bibitem[\protect\citeauthoryear{Wilson}{Wilson}{1972}]%
        {wilson1972}
\bibfield{author}{\bibinfo{person}{Dennis~L Wilson}.}
  \bibinfo{year}{1972}\natexlab{}.
\newblock \showarticletitle{Asymptotic properties of nearest neighbor rules
  using edited data sets}.
\newblock \bibinfo{journal}{\emph{Transactions on Systems, Man and
  Cybernetics}} (\bibinfo{year}{1972}).
\newblock


\bibitem[\protect\citeauthoryear{Xiong, Johnson, Xu, and Corso}{Xiong
  et~al\mbox{.}}{2012}]%
        {xiong2012random}
\bibfield{author}{\bibinfo{person}{Caiming Xiong}, \bibinfo{person}{David
  Johnson}, \bibinfo{person}{Ran Xu}, {and} \bibinfo{person}{Jason~J Corso}.}
  \bibinfo{year}{2012}\natexlab{}.
\newblock \showarticletitle{Random forests for metric learning with implicit
  pairwise position dependence}. In \bibinfo{booktitle}{\emph{KDD}}.
\newblock


\bibitem[\protect\citeauthoryear{Yeh, Kim, Yen, and Ravikumar}{Yeh
  et~al\mbox{.}}{2018}]%
        {yeh2018representer}
\bibfield{author}{\bibinfo{person}{Chih-Kuan Yeh}, \bibinfo{person}{Joon Kim},
  \bibinfo{person}{Ian En-Hsu Yen}, {and} \bibinfo{person}{Pradeep~K
  Ravikumar}.} \bibinfo{year}{2018}\natexlab{}.
\newblock \showarticletitle{Representer point selection for explaining deep
  neural networks}. In \bibinfo{booktitle}{\emph{NeurIPS}}.
\newblock


\bibitem[\protect\citeauthoryear{Zhao, Su, Ge, and Fan}{Zhao
  et~al\mbox{.}}{2016}]%
        {zhao2016propensity}
\bibfield{author}{\bibinfo{person}{Peng Zhao}, \bibinfo{person}{Xiaogang Su},
  \bibinfo{person}{Tingting Ge}, {and} \bibinfo{person}{Juanjuan Fan}.}
  \bibinfo{year}{2016}\natexlab{}.
\newblock \showarticletitle{Propensity score and proximity matching using
  random forest}.
\newblock \bibinfo{journal}{\emph{Contemporary Clinical Trials}}
  (\bibinfo{year}{2016}).
\newblock


\bibitem[\protect\citeauthoryear{Zhou, Zhou, Ning, Yang, and Li}{Zhou
  et~al\mbox{.}}{2015}]%
        {zhou2015two}
\bibfield{author}{\bibinfo{person}{Qi-Feng Zhou}, \bibinfo{person}{Hao Zhou},
  \bibinfo{person}{Yong-Peng Ning}, \bibinfo{person}{Fan Yang}, {and}
  \bibinfo{person}{Tao Li}.} \bibinfo{year}{2015}\natexlab{}.
\newblock \showarticletitle{Two approaches for novelty detection using random
  forest}.
\newblock \bibinfo{journal}{\emph{Expert Systems with Applications}}
  (\bibinfo{year}{2015}).
\newblock


\bibitem[\protect\citeauthoryear{Zhou, Zhou, and Hooker}{Zhou
  et~al\mbox{.}}{2018}]%
        {zhou2018approximation}
\bibfield{author}{\bibinfo{person}{Yichen Zhou}, \bibinfo{person}{Zhengze
  Zhou}, {and} \bibinfo{person}{Giles Hooker}.}
  \bibinfo{year}{2018}\natexlab{}.
\newblock \showarticletitle{Approximation trees: Statistical stability in model
  distillation}.
\newblock \bibinfo{journal}{\emph{arXiv preprint arXiv:1808.07573}}
  (\bibinfo{year}{2018}).
\newblock


\end{thebibliography}

\clearpage
\setcounter{table}{0}
\renewcommand{\thetable}{A\arabic{table}}
\setcounter{figure}{0}
\renewcommand\thefigure{A\arabic{figure}}

\appendix

\twocolumn[\section{User Study Materials}]

\begin{figure}[ht!]
\centering
    \fbox{\includegraphics[width=0.476\textwidth]{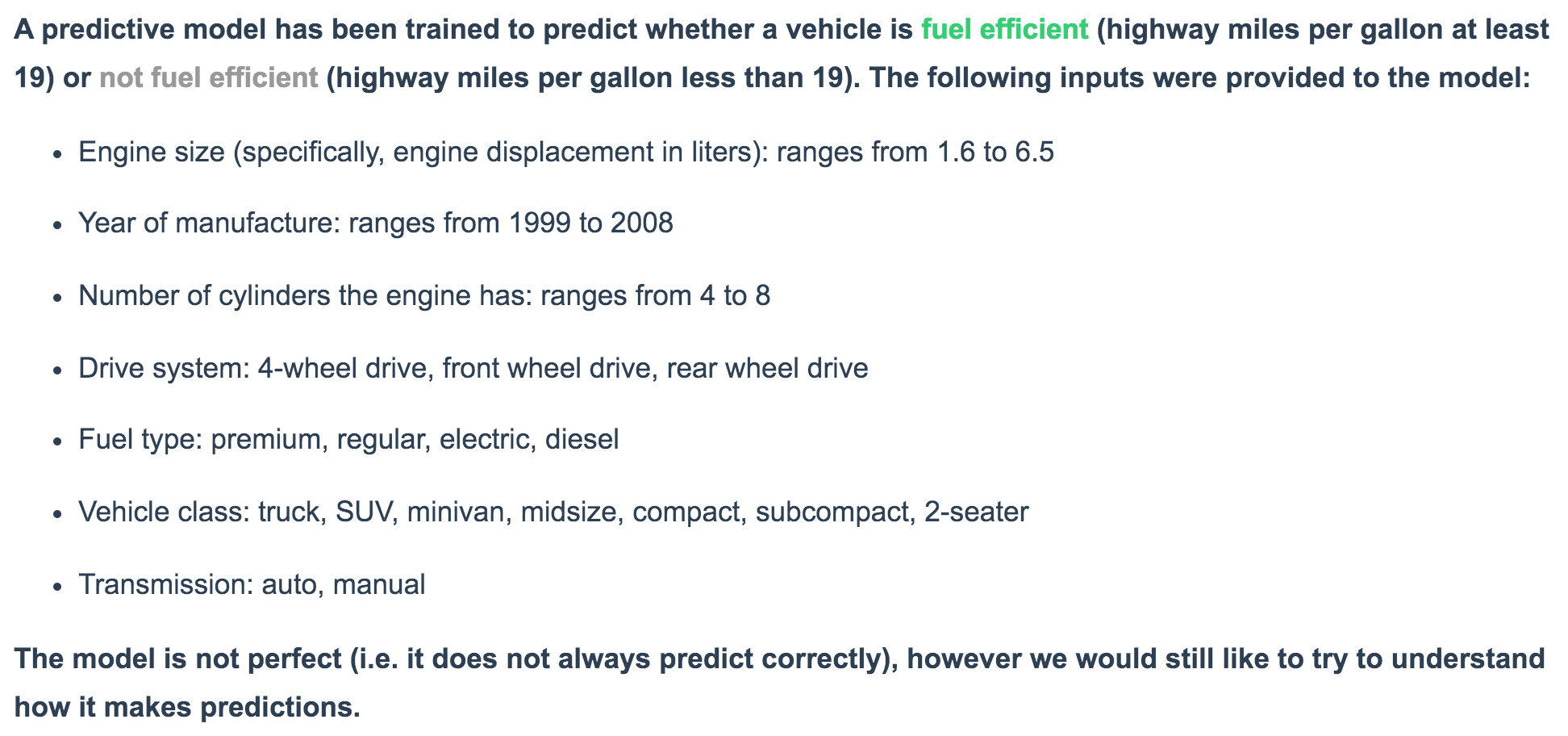}}
    \caption{Model inputs presented to users in both prototype and Shapley conditions.}
    \label{fig:userstudy_inputs}
\end{figure}

\begin{figure}[ht!]
\centering
    \fbox{\includegraphics[width=0.476\textwidth]{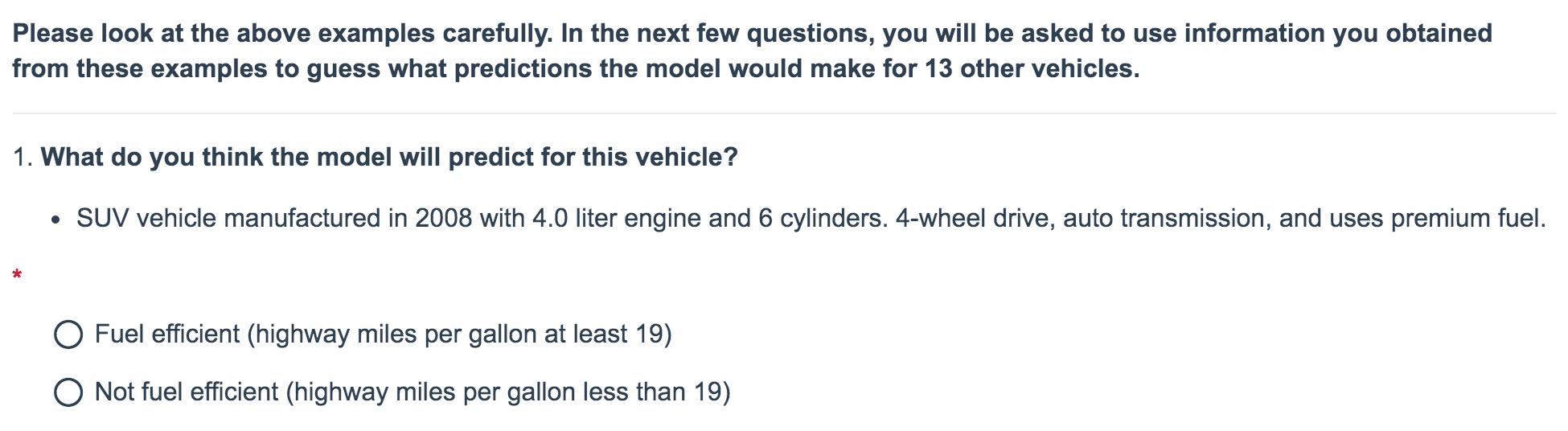}}
    \caption{An example question answered by users in both conditions. Each question represents a vehicle. Each user is asked to evaluate 13 such questions.}
    \label{fig:userstudy_questions}
\end{figure}

\begin{figure}[ht!]
\centering
    \fbox{\parbox{\linewidth}{
    \includegraphics[width=0.98\linewidth]{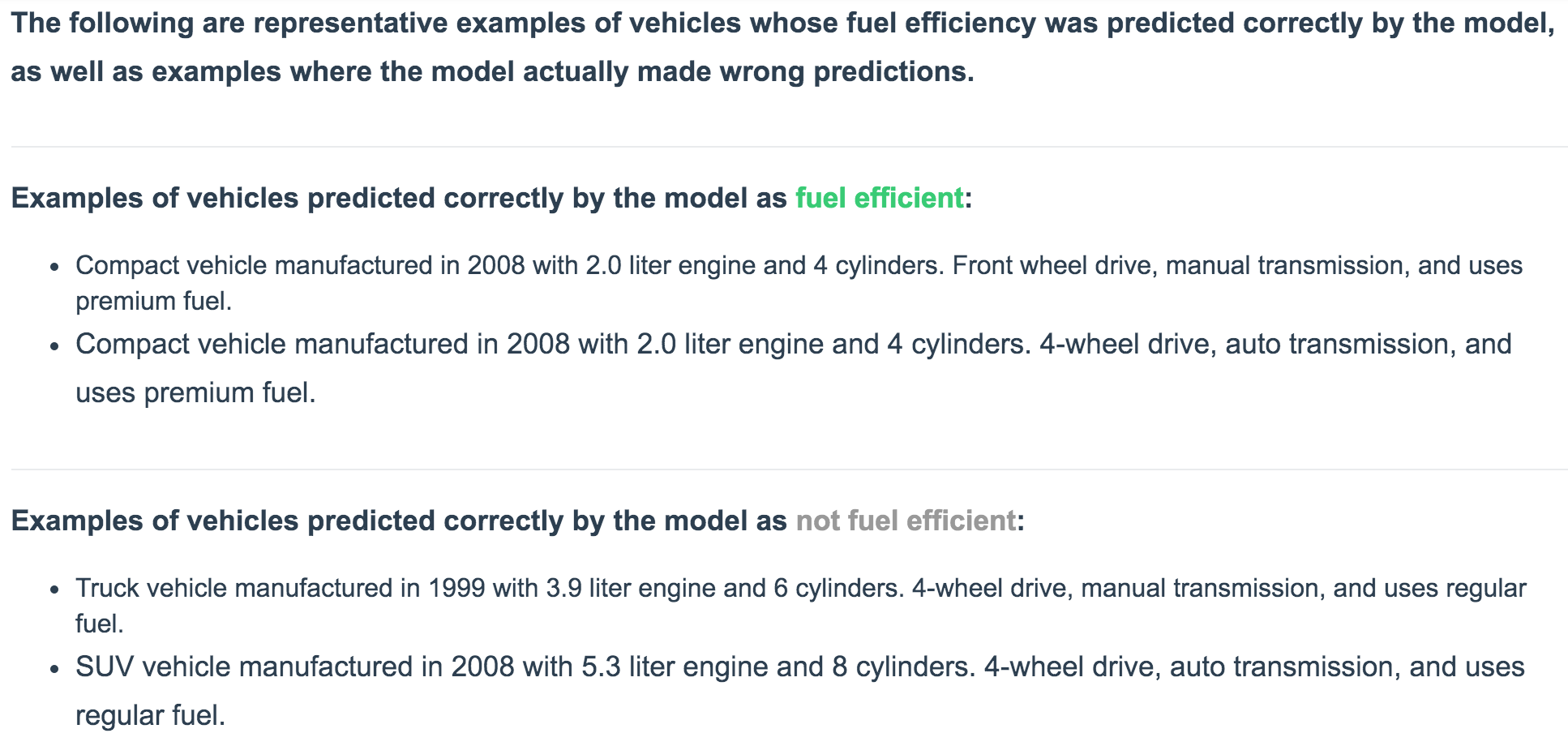}
    
    \includegraphics[width=0.98\linewidth]{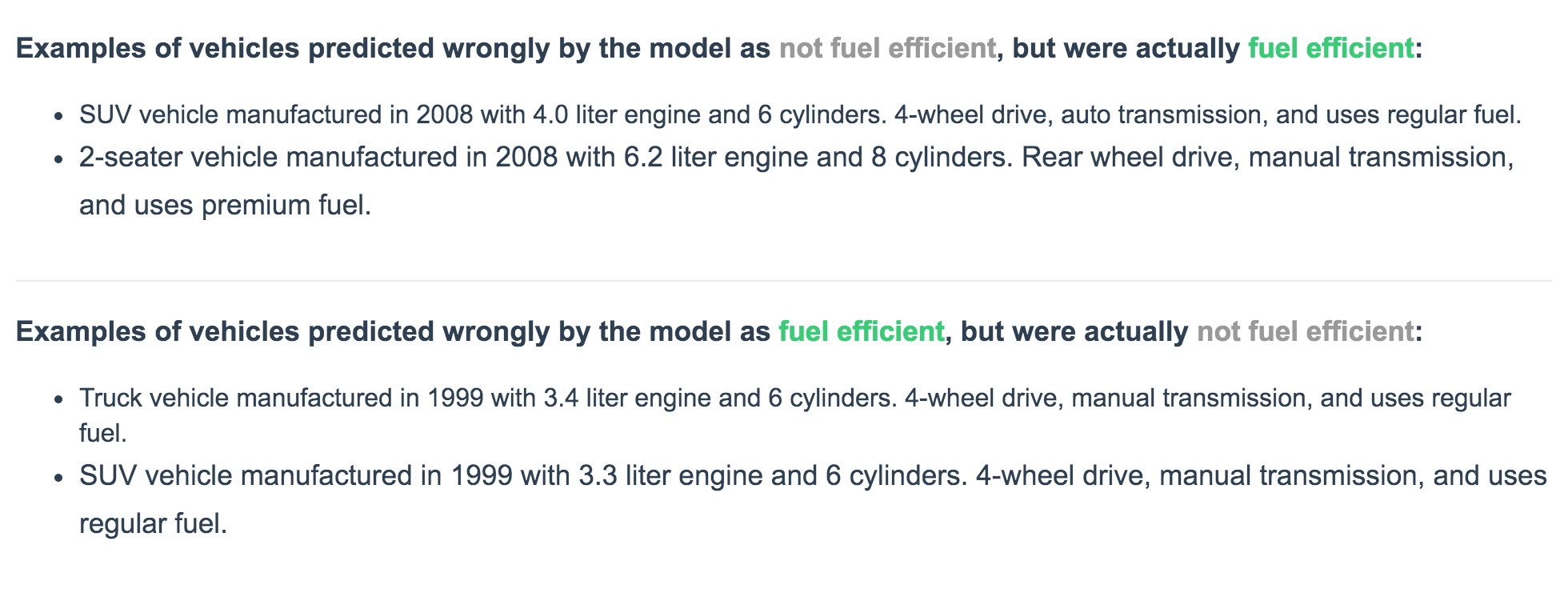}
    }}
    \caption{Prototypes presented to users in prototypes condition.}
    \label{fig:userstudy_prot}
\end{figure}


\begin{figure}
\centering
\fbox{\parbox{\linewidth}{
    \includegraphics[width=0.98\linewidth]{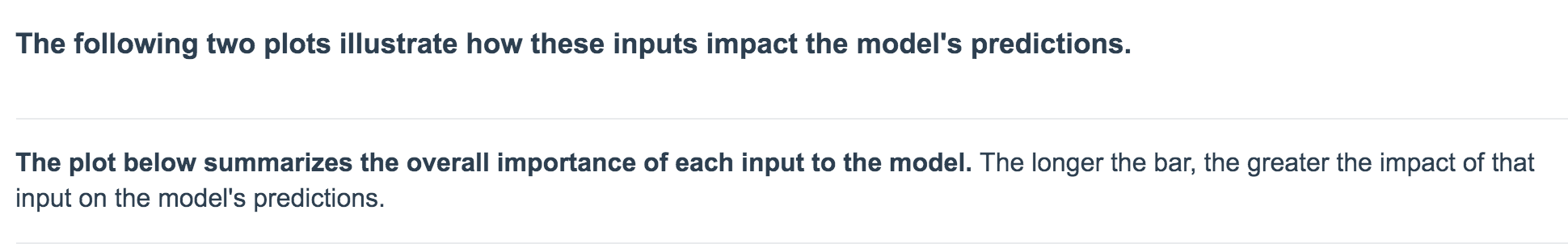}
    
    \includegraphics[width=0.98\linewidth]{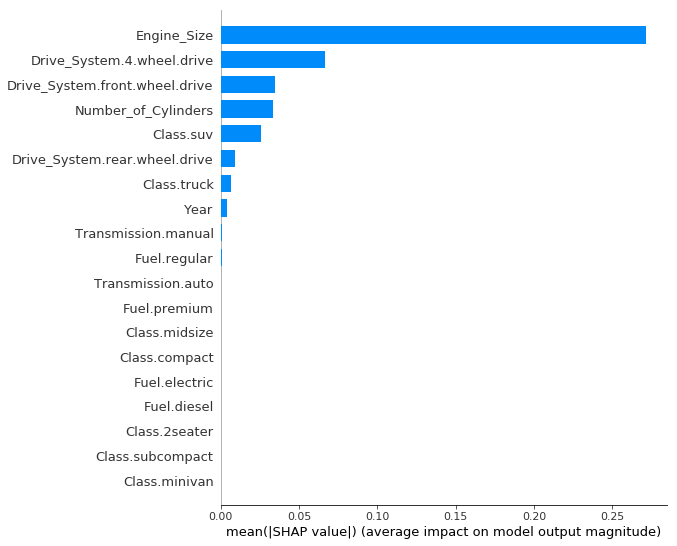}
    
     \includegraphics[width=0.98\linewidth]{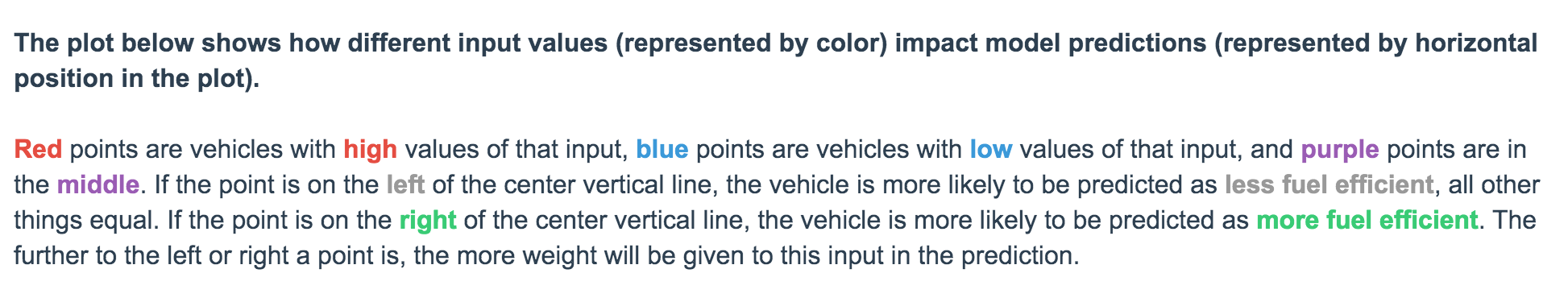}
     
      \includegraphics[width=0.98\linewidth]{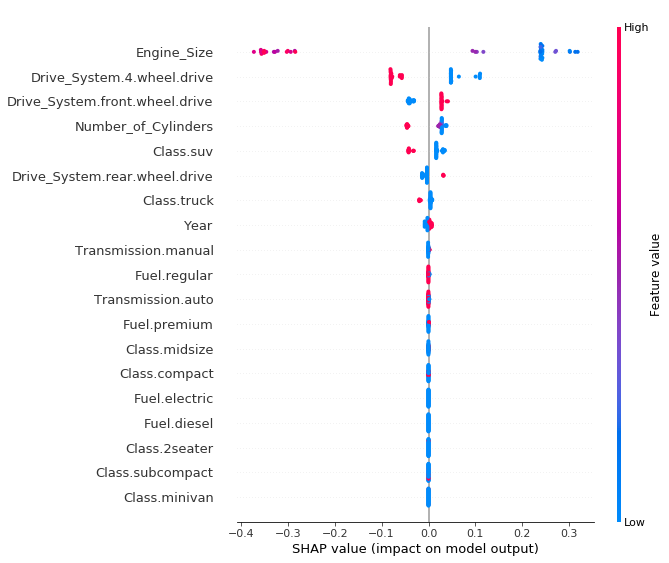}
      
       \includegraphics[width=0.98\linewidth]{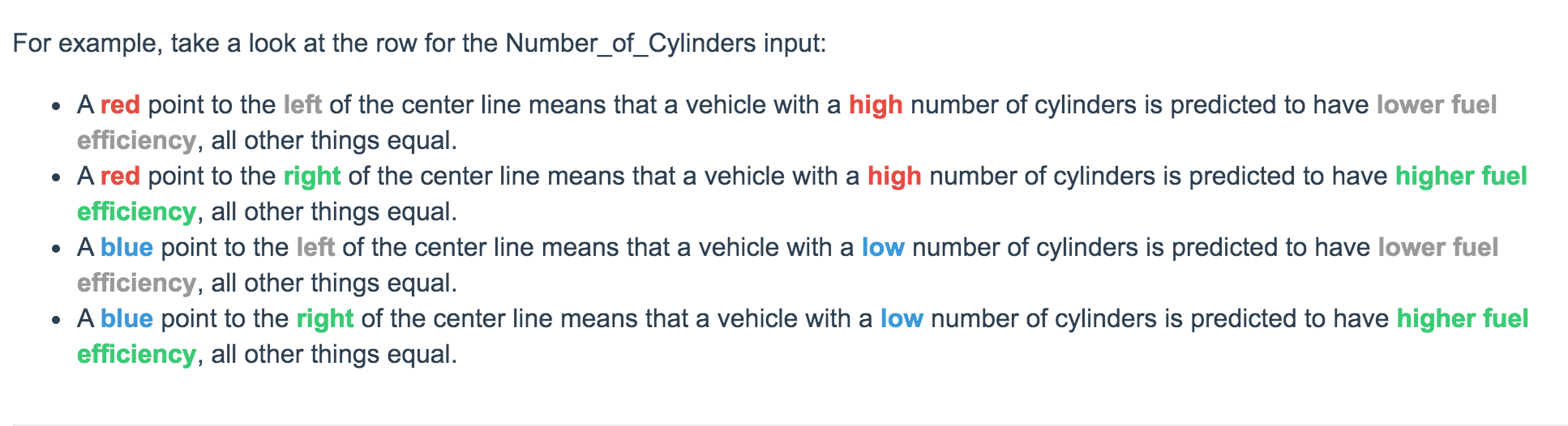}
}}
    \caption{Shapley values feature attribution plots presented to users in Shapley condition.}
    \label{fig:userstudy_shap}
\end{figure}

\section{Proof of Lemma 1}
\begin{lemmaproof} The objective function (\ref{eqn:objectivefixed}) is non-negative, monotone and submodular.
 \end{lemmaproof}
 \begin{proof}[Proof (Lemma 1)]
 Observe that whenever $X\subseteq Y$, we have $f(X) \geq f(Y)$, since adding more points to a set can only make the closest point to a given point closer. From this, monotonicity and non-negativity is immediate, since $f(P)\geq f(P\cup M)$.
 
 To establish submodularity, we will show that the function $f$ of (\ref{eqn:objective}) satisfies \begin{equation*} 
 f(Y)-f(Y\cup \{t\}) \leq f(X)-f(X\cup \{t\}) 
 \end{equation*} 
 whenever $X\subseteq Y\subseteq S$. The inequality of definition \ref{def:submod} then follows for $g$ by plugging into its definition (\ref{eqn:objectivefixed}).
 
 For any point $s\in S$, define $p_M(s)$ to be the closest point to $s$ in $M$ of the same class, that is,
 $$ p_M(s) = \underset{m\in M: c(m)=c(s)}{\arg\min} d(s,m). $$
 Then we can rewrite $f(M)$ as
 $$ \sum_{s\in S} d(s,p_M(s)), $$
 and it suffices to show that
 \begin{eqnarray*} & & d(s,p_Y(s))-d(s,p_{Y\cup \{t\}}(s)) \\
 &\leq& d(s,p_X(s)) - d(s,p_{X\cup \{t\}}(s)).
 \end{eqnarray*}
 for all $s\in S$. Both sides of this inequality are non-negative (+), since adding points can only shorten the distance to the closest point. Suppose $p_{Y\cup \{t\}}(s)\in Y$. Then it must be equal to $p_Y(s)$, since the closest point is present in $Y$, and so the first line is 0, and the inequality follows from (+).
 
 Suppose instead $p_{Y\cup\{t\}}(s)\not\in Y$. Then it must be $t$. So $p_{X\cup\{t\}}(s)=t$ as well (since $X\subseteq Y$), and the inequality reduces to $d(s,p_Y(s))\leq d(s,p_X(s))$. But this is immediate, since $Y\supseteq X$ and adding more points can only shorten the distance to the closest point. 
 \end{proof}
 
\end{document}